\documentclass[11pt]{article}
\usepackage{acl}
\usepackage{times}
\usepackage{latexsym}
\usepackage{amsmath}
\usepackage[T1]{fontenc}
\usepackage[utf8]{inputenc}
\usepackage{microtype}
\usepackage{inconsolata}
\usepackage{graphicx}
\usepackage{booktabs}
\usepackage{placeins}
\usepackage{stfloats}
\usepackage{colortbl}
\usepackage[most]{tcolorbox}
\definecolor{caseSource}{HTML}{28617A}
\definecolor{caseEdited}{HTML}{97511D}

\newtcolorbox{conclusionbox}{
    enhanced,
    colback=blue!3,
    colframe=blue!55!black,
    boxrule=0pt,
    leftrule=2.2pt,
    arc=1pt,
    left=7pt,
    right=7pt,
    top=6pt,
    bottom=6pt,
    before skip=8pt,
    after skip=8pt,
}
\title{Beyond the Opening: Tracing the First Token's Persistent Influence on Reasoning Length}

\author{
  \textbf{Ruochen Liu}\textsuperscript{1},
  \textbf{Hao Yan}\textsuperscript{2},
  \textbf{Ziluowen Luo}\textsuperscript{2},
  \textbf{Senzhang Wang}\textsuperscript{2},
  \textbf{Sihong Xie}\textsuperscript{1}
  \\
  \textsuperscript{1}The Hong Kong University of Science and Technology (Guangzhou) \\
  \textsuperscript{2}Central South University, China \\
  \texttt{\{csuyh1999, lzlwddl, szwang\}@csu.edu.cn} \\
  \texttt{rliu190@connect.hkust-gz.edu.cn, sihongxie@hkust-gz.edu.cn}
}

\begin{document}
\maketitle
\begin{abstract}

Understanding how individual token choices affect reasoning length is important for characterizing reasoning models. Building on prior findings that changing the first reasoning token (hereafter, the first token) can substantially alter reasoning length, we focus on this initial choice to investigate how its influence operates during subsequent generation. One possibility is that the first token changes the early text that follows, which in turn affects reasoning length. However, replacing only the first token of an existing reasoning prefix and resuming generation substantially changes the remaining reasoning length, even though all other prefix tokens are unchanged. To trace this persistent influence, we compare next-token distributions while varying only the first token and find that sensitivity is concentrated at a small set of generation positions. Blocking attention from these positions to the first token reduces the length gap between first-token conditions. Using next-token distributions from a different first-token condition at these positions also substantially changes subsequent reasoning length. Together, these findings suggest that attention to the first token at a few sensitive positions helps sustain its length effect, illustrating how a local token choice persistently influences a sequence-level outcome.
\end{abstract}

\section{Introduction}
\label{sec:intro}

Recent reasoning models, such as DeepSeek-R1~\citep{deepseek2025r1}, have demonstrated strong performance on complex tasks including mathematics and programming. These models solve problems by generating explicit chains of thought (CoTs) before producing final answers~\citep{wei2022chain}. However, the length of these reasoning traces can vary substantially across problems and even across different generations of the same problem~\citep{lin2025thinking,nguyen2026forks}, resulting in large differences in inference cost and potentially task performance~\citep{chen2024overthinking,muennighoff2025s1,aggarwal2025l1,snell2024scaling}. Understanding how reasoning length is determined during autoregressive generation is therefore important for characterizing the behavior of reasoning models and developing more effective ways to control their reasoning processes~\citep{arora2025efficient}.

Reasoning length is a property of the entire generated sequence, whereas generation proceeds through individual token choices. Understanding how these choices affect overall length is therefore an important aspect of understanding what determines reasoning length. Previous studies show that local token choices can substantially affect reasoning length~\citep{muennighoff2025s1,wang-etal-2025-wait}. In particular, studies of the first reasoning token (hereafter, the first token) show that its distribution is concentrated on a few candidate tokens associated with markedly different reasoning lengths, and that changing this token can alter subsequent reasoning length~\citep{lin2025thinking,nguyen2026forks}. However, the mechanisms through which local token choices affect subsequent token predictions and lead to differences in reasoning length remain insufficiently understood. The few prevalent first-token candidates and their substantial length effects offer a tractable setting for studying how this choice affects reasoning length through subsequent generation.

Intuitively, the first token may affect subsequent reasoning length by changing the early text generated after it. To test whether its length effect depends entirely on differences in this early text, we take a prefix from an existing reasoning trace and replace only its first token, keeping remaining prefix tokens unchanged. We then allow the model to continue from the modified prefix. Even with the remaining prefix tokens held fixed, substituting a first token typically associated with longer reasoning significantly increases the remaining reasoning length, whereas substituting one associated with shorter reasoning reduces it. 
These results show that early textual differences do not fully explain the first-token length effect, as the first token can still influence subsequent reasoning length when the intervening prefix text is held fixed.

To trace this persistent influence, we compare next-token distributions under different first-token conditions while holding all other prefix tokens fixed, examining sensitivity to the first token at each generation position. We find that these differences in next-token distributions are highly concentrated at a small set of positions, which also exhibit stronger attention to the first token. We perform two interventions to test whether these positions contribute to the length effect. Blocking their attention to the first token substantially reduces the reasoning-length gap between first-token conditions. 
Using next-token distributions from a different first-token condition at these positions substantially changes subsequent reasoning length, even when the original first token is retained. Together, these results suggest that attention to the first token at a few sensitive positions helps sustain its influence on subsequent predictions and reasoning length.
More broadly, this study links a local token choice to overall reasoning length through its persistent influence on subsequent token predictions. Our key findings are threefold. 
\begin{itemize}
\item The first-token length effect persists beyond the intervening reasoning text. Even when the subsequent reasoning prefix is held fixed, changing the first token substantially alters the remaining reasoning length.

\item The first token's persistent influence on subsequent predictions is concentrated at a small set of generation positions, which also exhibit stronger attention to the first token.

\item Sensitive positions contribute to sustaining the first-token length effect. Blocking their attention to the first token reduces the length gap, while replacing their next-token distributions redirects reasoning length in both directions.
\end{itemize}

\section{Related Work}
\label{sec:related}
\paragraph{Local token choices and reasoning length.}
Interventions on token choices during generation can substantially change reasoning length. \citet{muennighoff2025s1} extend reasoning through budget forcing, suppressing termination and appending \texttt{Wait} when the model attempts to stop. Conversely, \citet{wang-etal-2025-wait} shorten reasoning by suppressing reflection-related tokens such as \texttt{Wait} and \texttt{Hmm} during decoding. At the start of reasoning, \citet{lin2025thinking} identify first tokens associated with fast and slow thinking and show that forcing a short-associated first token reduces reasoning length. \citet{nguyen2026forks} further connect initial token choices to linear and backtracking reasoning modes with different lengths and accuracies. These studies demonstrate that local token choices can alter reasoning length. We build on this work by examining how the first token's influence on length is carried through subsequent generation.

\paragraph{Internal representations and control of reasoning length.}
Reasoning length can be controlled through budget forcing~\citep{muennighoff2025s1}, reinforcement learning with length constraints~\citep{aggarwal2025l1}, and training on compressed reasoning traces~\citep{xia2025tokenskip}. Complementary work studies internal representations. \citet{sheng2025planning} predict reasoning length from question activations and identify a direction that modulates reasoning strength. Activation steering controls thinking speed~\citep{lin2025thinking} and behaviors such as backtracking and uncertainty~\citep{venhoff2025steering}. ThinkEdit edits attention-head output projections to mitigate overly short reasoning~\citep{sun2025thinkedit}, while FlowSteer learns a nonlinear transformation from verbose to concise reasoning representations~\citep{li2026flowsteer}. These approaches demonstrate ways to regulate reasoning length.

\section{Preliminaries}
\label{sec:preliminaries}
\subsection{Problem Formulation}
\label{sec:formulation}

Given a problem $x$, a reasoning model generates a chain of thought $r=(r_1,\ldots,r_L)$ followed by a final answer. We define reasoning length as $L=|r|$, excluding reasoning delimiters and the final answer, and denote the first reasoning token by $z=r_1$. Our goal is to understand how the influence of $z$ on reasoning length persists as generation unfolds, and through which later generation states it operates.

\subsection{First-Token Effects on Reasoning Length}
\label{sec:first-token}

\begin{figure}[t]
    \centering
    \includegraphics[width=\columnwidth]{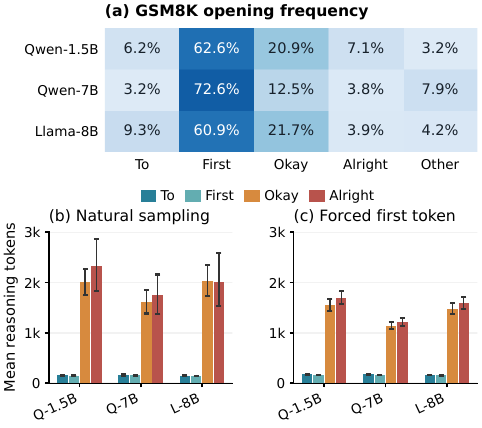}
    \caption{First-token effects on GSM8K: (a) first token frequencies under natural sampling; (b) mean reasoning length conditioned on the natural first token; (c) mean reasoning length under forced first tokens. Heatmap cells report percentages, bar colors identify first tokens. Other includes all remaining first tokens. Q and L denote Qwen and Llama. Error bars indicate 95\% bootstrap confidence intervals.}
    \label{fig:preliminary}
\end{figure}

We illustrate the first-token effect on reasoning length using GSM8K. Under natural sampling, the first reasoning token is concentrated in a small set of first tokens, primarily \texttt{Okay}, \texttt{Alright}, \texttt{To}, and \texttt{First} (Figure~\ref{fig:preliminary}a). Reasoning traces beginning with \texttt{Okay} or \texttt{Alright} are substantially longer on average than those beginning with \texttt{To} or \texttt{First} (Figure~\ref{fig:preliminary}b). This length contrast persists when we force the reasoning trace for each question to begin with each of the four tokens, indicating that the first token can shape subsequent reasoning length rather than merely correlate with it (Figure~\ref{fig:preliminary}c). Further experimental details and numerical results are provided in Appendix~\ref{app:baselines}. These observations motivate our central question: how does the first token influence reasoning length as reasoning unfolds?

\section{Experimental Setup}
\label{sec:setup}
\paragraph{Models and tasks.}
We study three DeepSeek-R1 distilled models~\citep{deepseek2025r1}: Distill-Qwen-1.5B, Distill-Qwen-7B, and Distill-Llama-8B, abbreviated as Qwen-1.5B, Qwen-7B, and Llama-8B. We use all 500 MATH-500 test questions~\citep{hendrycks2021math,lightman2023verify} and a fixed random sample of 500 GSM8K training questions~\citep{cobbe2021gsm8k}, with the same question sets across models and conditions. HumanEval~\citep{chen2021humaneval} and FOLIO~\citep{han2024folio} provide coding and logical reasoning tasks for the subsequent generalization analysis.

\paragraph{Reference trajectories.}
We use two types of reference trajectories to study first-token effects on reasoning length. Naturally sampled trajectories characterize first-token frequencies and their association with reasoning length. We focus on \texttt{Okay}, \texttt{Alright}, \texttt{First}, and \texttt{To}, which account for most naturally occurring firsts across our evaluated models and datasets (Appendix~\ref{app:baselines}). Forced first token trajectories begin with one of these long-associated (\texttt{Okay}, \texttt{Alright}) or short-associated (\texttt{First}, \texttt{To}) tokens, followed by freely sampled continuation. They serve as unmodified references for subsequent intervention experiments, with comparisons matched by question.

\paragraph{Sampling and evaluation.}
We draw three samples per question and condition at temperature $0.6$ and top-$p=0.95$. Reasoning lengths are measured within an 8,192-token observation horizon; unfinished traces are retained at their observed lengths. We report total reasoning length for free generation and remaining length for prefix-preserving interventions. Additional details are provided in Appendix~\ref{app:setup-details}.

\section{How the First Token Shapes Reasoning Length}
\label{sec:experiments}
To understand how the first token shapes subsequent reasoning length, we first use prefix-preserving interventions to examine whether its influence persists after a reasoning path has formed. We then identify later generation states whose predictions are sensitive to the first-token condition. Finally, we use attention masking and distribution replacement to test whether direct reads of the first token at these states sustain length differences and whether changing their generation choices can reproduce the length effect.

\subsection{Persistent First-Token Influence}
\label{sec:prefix}
\begin{figure*}[t]
    \centering
    \includegraphics[width=\textwidth]{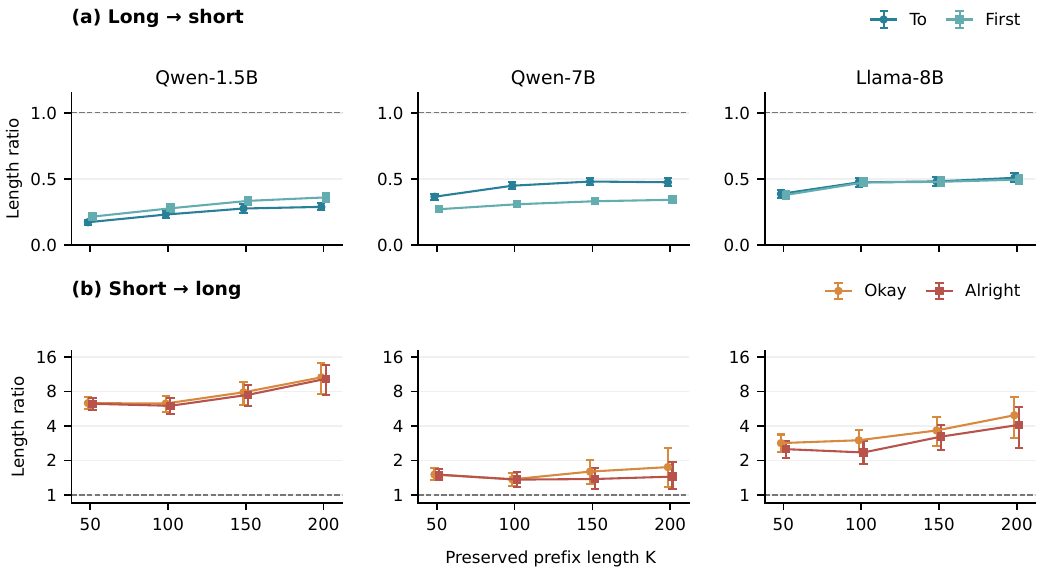}
    \caption{First-token replacement on GSM8K with subsequent prefix tokens held fixed. (a) Long-to-short replacements. (b) Short-to-long replacements. Values are ratios of mean remaining reasoning lengths to matched unchanged-prefix references, averaging the observed source suffix and two new continuations. Error bars indicate 95\% bootstrap confidence intervals.}
    \label{fig:prefix-persistence}
\end{figure*}

\paragraph{Setup.} To test whether the first-token effect persists beyond the initial selection of a reasoning path, we replace the first token of a forced first token trajectory while holding the subsequent prefix tokens fixed. For prefix lengths \(K\in\{50,100,150,200\}\), we replace long-associated first token (\texttt{Okay} or \texttt{Alright}) with short-associated first token (\texttt{To} or \texttt{First}), and vice versa:
\begin{equation}
r'_{1:K}=(z',r_2,\ldots,r_K).
\end{equation}
We sample three continuations from each edited prefix and compare their mean remaining reasoning length with an unchanged-prefix reference, averaging the observed source suffix \(r_{K+1:L}\) and two newly sampled continuations. Further experimental details and results are provided in Appendix~\ref{app:prefix}.

\paragraph{The length effect extends beyond the initial text.}
Figure~\ref{fig:prefix-persistence} shows that changing the first token shifts the remaining reasoning length toward the tendency associated with the replacement first token. Specifically, replacing a long-associated first token with a short-associated one shortens the subsequent reasoning, while the reverse replacement lengthens it. If the first token affected reasoning length only by inducing different early text, preserving that text should eliminate its influence on subsequent generation. However, the effect remains evident even when the following 199 tokens are preserved. These results indicate that the first token's influence on reasoning length cannot be explained solely by the early text it induces, and persists even when the intervening text is held fixed.

\begin{table}[t]
\centering\small
\setlength{\tabcolsep}{4pt}
\begin{tabular}{llrr}
\toprule
Model & Opening & Rem. tokens & Acc. (\%) \\
\midrule
\rowcolor{black!10}\multicolumn{4}{c}{Long-to-short} \\
Qwen-1.5B & Source & 1,461 & 88.6 \\
 & \texttt{To} & 419 & 86.3 \\
 & \texttt{First} & 525 & 86.9 \\
\midrule
Qwen-7B & Source & 1,019 & 94.8 \\
 & \texttt{To} & 483 & 93.7 \\
 & \texttt{First} & 347 & 94.5 \\
\midrule
Llama-8B & Source & 1,348 & 89.0 \\
 & \texttt{To} & 683 & 86.6 \\
 & \texttt{First} & 666 & 85.9 \\
\rowcolor{black!10}\multicolumn{4}{c}{Short-to-long} \\
Qwen-1.5B & Source & 48 & 74.1 \\
 & \texttt{Okay} & 510 & 75.4 \\
 & \texttt{Alright} & 496 & 77.4 \\
\midrule
Qwen-7B & Source & 43 & 82.8 \\
 & \texttt{Okay} & 75 & 83.0 \\
 & \texttt{Alright} & 62 & 82.4 \\
\midrule
Llama-8B & Source & 47 & 54.5 \\
 & \texttt{Okay} & 235 & 56.3 \\
 & \texttt{Alright} & 194 & 55.4 \\
\bottomrule
\end{tabular}
\caption{Remaining reasoning length and answer accuracy on GSM8K at $K=200$. }
\label{tab:prefix-accuracy}
\end{table}

\paragraph{Edited prefixes remain compatible with effective question answering.}
Because the preserved prefix text was generated under the original first token, replacing that token may disrupt reasoning and thereby alter length. However, Table~\ref{tab:prefix-accuracy} shows that shortened continuations still answer most questions correctly, though less accurately than the matched source trajectories, while lengthened continuations generally maintain or improve mean accuracy. These results suggest that the length changes do not simply reflect a general breakdown in answering ability caused by prefix editing.

\begin{conclusionbox}
    \textbf{Conclusion.}
These results show that changing the first token alters subsequent reasoning length even with the following prefix fixed, while preserving effective question answering. This provides a controlled setting in which the first-token condition is varied while the subsequent prefix text remains unchanged. Building on this setting, we next investigate what changes in subsequent generation when the first token is replaced, and how these changes affect reasoning length.
\end{conclusionbox}

\subsection{First-token Sensitivity Concentrates at Few Positions}
\label{sec:sensitive-states}
\paragraph{Setup.} Since reasoning unfolds through successive next-token predictions, we investigate the first token's influence on subsequent generation by examining how replacing it changes these predictive distributions. Using the controlled setting established in the previous section, we compare predictive distributions under different first-token conditions while keeping the subsequent reasoning prefix fixed. At position $t$, we define
\begin{equation}
p_t^{z}=P_\theta(\cdot\mid x,z,r_2,\ldots,r_t),
\end{equation}
and quantify the sensitivity of the next-token prediction to the first-token condition using Jensen--Shannon (JS) divergence~\citep{lin1991divergence}. For a long-associated first token \(z_{\mathrm{long}}\) and a
short-associated first token \(z_{\mathrm{short}}\), we define:
\begin{equation}
\begin{aligned}
J_t &= \tfrac{1}{2}D_{\mathrm{KL}}\!\left(p_t^{z_{\mathrm{short}}}\middle\|m_t\right) \\
    &\quad + \tfrac{1}{2}D_{\mathrm{KL}}\!\left(p_t^{z_{\mathrm{long}}}\middle\|m_t\right), \\
m_t &= \tfrac{1}{2}\left(p_t^{z_{\mathrm{short}}}+p_t^{z_{\mathrm{long}}}\right).
\end{aligned}
\end{equation}
 We present \texttt{First}--\texttt{Okay} as a representative long--short pair; results for the other three pairs are provided in Appendix~\ref{app:sensitive-states}.

\begin{figure}[t]
    \centering
    \includegraphics[width=\linewidth]{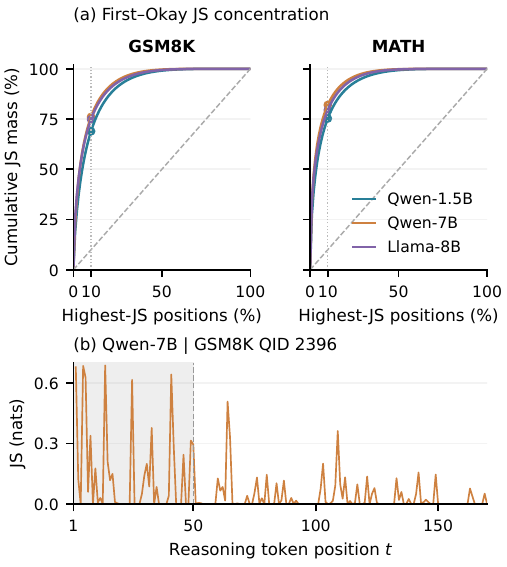}
    \caption{Concentrated first-token sensitivity. (a) Aggregate JS concentration on GSM8K and MATH for all three models. Positions are ranked within each trajectory; curves show mean cumulative JS shares. Markers indicate the highest-JS 10\% of positions; diagonals denote uniform contributions. (b) Unsmoothed JS along one Qwen-7B GSM8K trajectory, illustrating intermittent peaks.}
    \label{fig:js-concentration}
\end{figure}

\paragraph{Sensitivity is concentrated at a small fraction of positions.}
Figure~\ref{fig:js-concentration}(a) shows that, across models and datasets, the highest-JS 10\% of positions account for approximately 69--82\% of total JS. Sensitivity to the first-token condition is therefore highly concentrated along reasoning trajectories, with a small fraction of positions accounting for most of the difference in predictive distributions. Figure~\ref{fig:js-concentration}(b) illustrates this pattern in an example trajectory. JS remains relatively low at most positions, interspersed with distinct peaks. Although these peaks are denser and larger early in generation, they also occur later, showing that high sensitivity is not confined to the initial reasoning stage. Additional concentration analyses are provided in Appendix~\ref{app:sensitive-states}.

\begin{figure}[t]
    \centering
    \includegraphics[width=\linewidth]{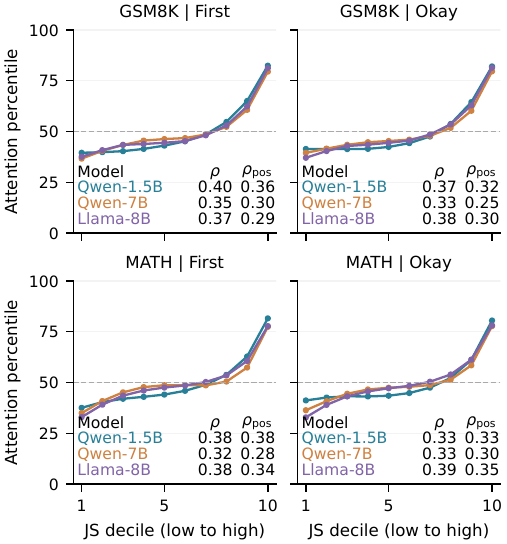}
    \caption{First-token attention versus JS on GSM8K (top) and MATH (bottom), under First (left) and Okay (right) first tokens. Curves show mean attention percentiles by within-trajectory JS decile. Insets report mean Spearman correlations before ($\rho$) and after ($\rho_{\mathrm{pos}}$) adjusting for token position.}
    \label{fig:attention-js}
\end{figure}

\paragraph{High-JS positions are more strongly associated with the first token.}
We further examine attention to the first token to assess whether high-JS positions are more strongly associated with it. Figure~\ref{fig:attention-js} shows that, across models, datasets, and first token conditions, positions with higher JS tend to assign greater attention to the first token. In the highest JS decile, first-token attention lies at the 77th--82nd within-trajectory percentile on average, and the mean within-trajectory Spearman correlation between JS and first-token attention ranges from 0.32 to 0.40. Since both quantities may vary systematically with generation position, we remove their linear position trends and find that the correlation remains positive, ranging from 0.25 to 0.38. These results indicate that high-JS positions not only exhibit greater predictive sensitivity to the first-token condition, but are also more strongly associated with the first token itself.

\begin{conclusionbox}
\textbf{Conclusion.}
    Taken together, these observations show that prediction differences induced by the first-token condition are highly concentrated at a small fraction of positions, and that these positions are more strongly associated with the first token itself. Based on these observations, we hypothesize that the first token may shape the eventual reasoning length primarily by influencing a small set of highly sensitive positions during reasoning. We next test this hypothesis experimentally.
\end{conclusionbox}

\subsection{Testing the Role of Sensitive Positions in Reasoning Length}
\label{sec:interventions}
\label{sec:attention}
\label{sec:distribution}
\begin{figure*}[t]
    \centering
    \includegraphics[width=\textwidth]{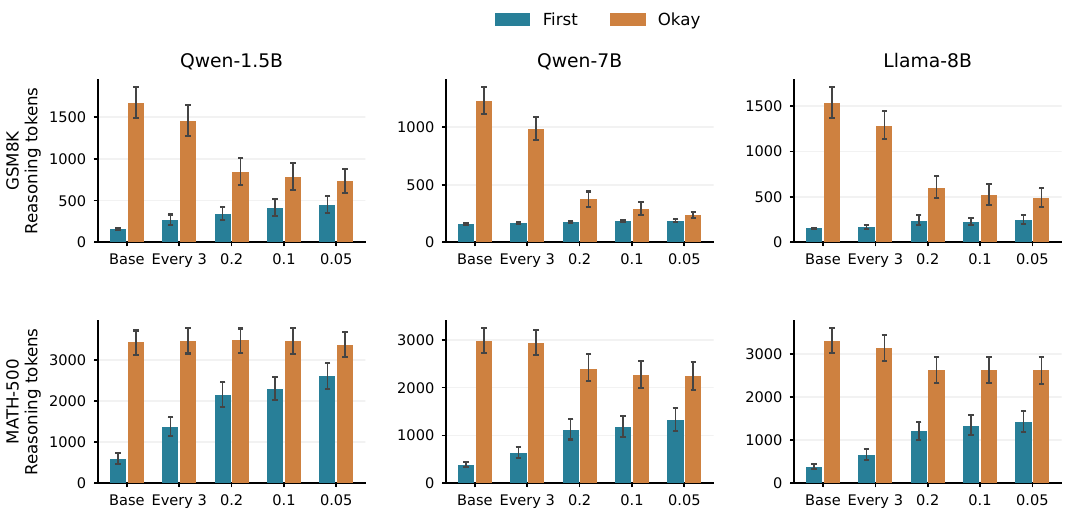}
    \caption{Reasoning length under first-token attention blocking on GSM8K (top) and MATH (bottom). Base denotes no intervention; Every 3 masks every third eligible position. The remaining groups mask positions with JS above the indicated threshold. Error bars indicate 95\% bootstrap confidence intervals.}
    \label{fig:attention-interventions}
\end{figure*}

\paragraph{Setup.} 
We perform two interventions to test whether sensitive positions contribute to the first-token length effect. \textit{Attention blocking} masks attention to the first reasoning token whenever $J_t>\tau$. For every layer $\ell$ and head $h$, we set the corresponding pre-softmax attention logit to
\begin{equation}
\widetilde{s}_{t,1}^{(\ell,h)}=
\begin{cases}
-\infty, & J_t>\tau,\\
s_{t,1}^{(\ell,h)}, & \text{otherwise},
\end{cases}
\end{equation}
while leaving all other logits unchanged before softmax. This tests whether direct attention to the first token at sensitive positions contributes to the length effect. \textit{Distribution replacement} instead uses
\begin{equation}
\widetilde{p}_t=
\begin{cases}
p_t^{z'}, & J_t>\tau,\\
p_t^{z}, & \text{otherwise},
\end{cases}
\end{equation}
where $z$ and $z'$ denote the source and alternative first tokens, respectively. Both distributions condition on the same subsequent reasoning prefix. We sample the next token from $\widetilde{p}_t$ to test whether changing predictions at sensitive positions shifts subsequent reasoning length. For both interventions, we focus on \texttt{First} and \texttt{Okay} as frequent short- and long-associated first tokens, respectively, using the same JS contrast as in Section~\ref{sec:sensitive-states}.

We use absolute thresholds because JS distributions differ across first token conditions and reasoning lengths. Selecting a fixed percentage would assign more intervention positions to longer chains, potentially including positions with low sensitivity. 
We evaluate both interventions at JS thresholds
$\tau\in\{0.05,0.1,0.2\}$.


\begin{table}[t]
\centering\small
\setlength{\tabcolsep}{3pt}
\begin{tabular}{llrrrr}
\toprule
Model & Opening & Every 3 & $0.2$ & $0.1$ & $0.05$ \\
\midrule
\rowcolor[gray]{0.92}\multicolumn{6}{c}{GSM8K} \\
Qwen-1.5B & First & 33.3 & 6.1 & 10.9 & 16.9 \\
 & Okay & 33.4 & 5.2 & 9.7 & 15.7 \\
Qwen-7B & First & 33.4 & 6.2 & 10.4 & 16.1 \\
 & Okay & 33.4 & 6.3 & 10.9 & 16.6 \\
Llama-8B & First & 33.3 & 9.5 & 16.2 & 23.2 \\
 & Okay & 33.4 & 8.5 & 14.9 & 21.4 \\
\rowcolor[gray]{0.92}\multicolumn{6}{c}{MATH-500} \\
Qwen-1.5B & First & 33.3 & 2.1 & 3.9 & 6.7 \\
 & Okay & 33.3 & 0.6 & 1.6 & 3.9 \\
Qwen-7B & First & 33.3 & 4.9 & 8.1 & 12.0 \\
 & Okay & 33.3 & 2.6 & 5.5 & 8.9 \\
Llama-8B & First & 33.3 & 5.3 & 9.5 & 13.9 \\
 & Okay & 33.3 & 2.9 & 5.5 & 8.6 \\
\bottomrule
\end{tabular}
\caption{Attention intervention rates (\%). Values are mean per-trajectory percentages of eligible positions. Every 3 denotes periodic masking; numeric headings denote JS thresholds.}
\label{tab:attention-intervention-rates}
\end{table}

\paragraph{First-token attention at sensitive positions sustains the length effect.}

Figure~\ref{fig:attention-interventions} shows that blocking attention to the first token at selected positions reduces the length gap between \texttt{First} and \texttt{Okay} by 51--96\% relative to the unmodified baselines across all three models and both datasets. This reduction reflects shorter \texttt{Okay} continuations, longer \texttt{First} continuations, or both. On Qwen-1.5B/MATH, the change is driven mainly by lengthening \texttt{First}, with relatively little change in \texttt{Okay}. The intervention thus weakens the distinction between first token conditions rather than uniformly shortening reasoning, although a residual gap remains. In every model--dataset combination, JS-based selection reduces the gap more than periodic masking while intervening at a smaller fraction of positions (Table~\ref{tab:attention-intervention-rates}), supporting the importance of where attention is blocked.

\begin{figure*}[t]
    \centering
    \includegraphics[width=\textwidth]{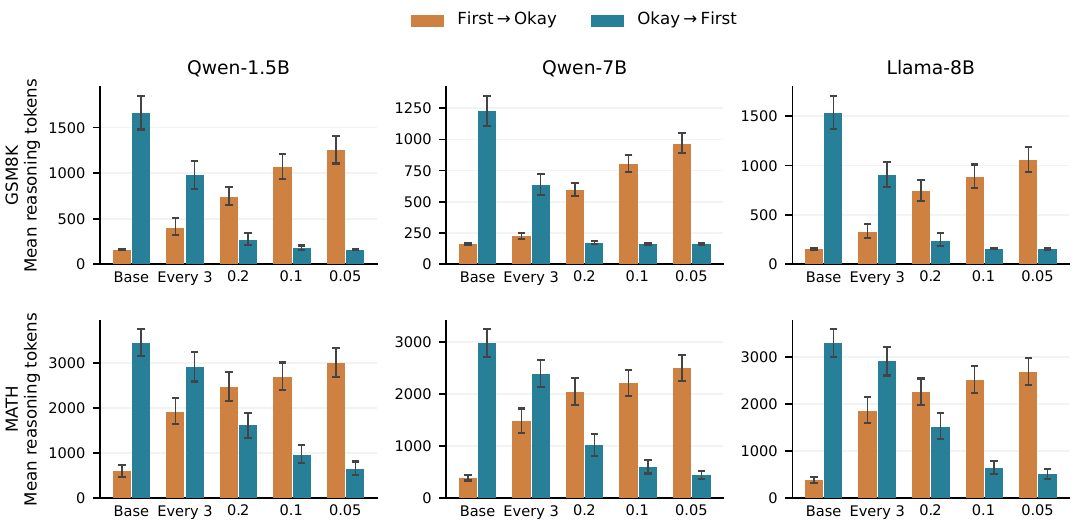}
    \caption{Reasoning length under distribution replacement on GSM8K (top) and MATH (bottom). Base denotes no intervention; Every 3 replaces distributions at every third eligible position; numeric headings denote JS thresholds. Error bars show 95\% question-bootstrap intervals.}
    \label{fig:distribution-interventions}
\end{figure*}

\paragraph{Distribution replacement redirects reasoning length.}
Figure~\ref{fig:distribution-interventions} shows bidirectional length shifts across all three models and both datasets. Replacing distributions from \texttt{First} with those from \texttt{Okay} lengthens reasoning, whereas the reverse replacement shortens it. Across models, datasets, and tested thresholds, First-to-Okay replacement increases mean length to 3.7--7.9 times its source baseline, while Okay-to-First replacement reduces it by 53--90\%. These shifts occur while retaining the source first token and changing only the sampling distributions at selected steps. At every tested threshold, JS-based replacement produces larger mean shifts toward the alternative first token than periodic replacement, despite a lower intervention rate (Table~\ref{tab:distribution-intervention-rates}). Thus, changing predictions within the candidate set can redirect subsequent reasoning length without changing the first token itself.

\paragraph{Targeting sensitive positions is more effective than broad periodic intervention.}
For both attention blocking and distribution replacement, JS-based selection produces larger changes in the first token dependent length difference than periodic intervention despite modifying a smaller fraction of positions (Tables~\ref{tab:attention-intervention-rates} and~\ref{tab:distribution-intervention-rates}). This consistent advantage across two distinct interventions shows that their effects cannot be explained simply by the number of modified generation steps. Instead, where the intervention is applied matters: positions that are more sensitive to the first-token condition exert a stronger aggregate influence on the resulting length difference.

\begin{conclusionbox}
    \textbf{Conclusion.} Together, these interventions indicate that attention to the first token at sensitive positions helps sustain its effect on reasoning length. Changing next-token distributions at these positions can redirect subsequent reasoning length even when the first token remains unchanged, linking its persistent influence on token predictions to overall length differences.
\end{conclusionbox}



\section{Discussion}
\label{sec:discussion}

The first-token effect raises a broader question about how local choices shape generation as a whole. A token need not explicitly specify a desired length to influence how much reasoning follows. Our results offer a starting point for considering how such connections arise and whether they extend to other tokens.

\paragraph{Local expressions and global generation properties.}
first tokens such as \texttt{First} and \texttt{Okay} do not explicitly specify reasoning length, yet they exhibit distinct length tendencies whose influence persists beyond the early text they induce. One possible explanation is that models learn associations between local expressions and sequence-level characteristics during training. For example, an expression that frequently co-occurs with longer reasoning may favor more extended continuations without conveying an explicit length instruction. From this perspective, a token's influence may reflect both its literal meaning and learned associations with broader generation patterns. Examining these associations may help explain how small differences in local expression relate to substantial differences in sequence-level behavior.

\paragraph{The first token as an entry point.}
Similar persistent influences may occur throughout generation, with the first token providing a tractable setting for observation and intervention. Its position is aligned across trajectories, and its values are concentrated in a small set of common first tokens. These properties enable consistent substitutions and allow us to trace their effects on subsequent predictions and overall length. Tokens at intermediate positions may play similar roles, although their identities and functions depend more strongly on the surrounding context. Studying the first token therefore offers a research strategy that can be explored more broadly: introduce controlled local substitutions, identify sensitive positions in subsequent generation, and intervene to examine how these dependencies contribute to global generation properties.

\paragraph{Future work.}
A next step is to investigate how the contributions of individual positions within the candidate set combine to shape reasoning length. Finer-grained interventions could distinguish positions with substantial effects on length from those that primarily alter local wording, and determine whether their contributions are independent, redundant, or synergistic. Analyses of the surrounding context and internal representations could further reveal which generation states make earlier cues influential and how predictive changes at these positions alter the subsequent course of reasoning. This would extend the set-level intervention evidence toward a more specific account of how local dependencies contribute to global generation properties.

\section{Conclusion}
\label{sec:conclusion}
We investigate how the first reasoning token influences subsequent reasoning length. Its effect persists when the intervening text is held fixed, while predictive sensitivity is concentrated at a small fraction of positions. Attention blocking and distribution replacement at these positions respectively attenuate length differences and redirect reasoning length. Together, these findings support a role for the selected candidate set in sustaining the first-token length effect.

\section*{Limitations}
\paragraph{Model scope.}
All three models studied in this work belong to the DeepSeek-R1 distilled series. Although they differ in size and backbone architecture, they share the same source model for distillation. Whether persistent first-token influence and concentrated sensitivity also occur in other model families and under different training procedures remains to be established.

\paragraph{From sensitive positions to reasoning length.}
We characterize the relationship between the first token and subsequent sensitive positions, and establish their contribution to the length effect through set-level interventions. This evidence concerns the selected set as a whole, leaving the contributions of individual positions unresolved. Finer-grained interventions are needed to determine which positions within the set drive the observed length changes.


\bibliography{references}

\clearpage
\appendix
\raggedbottom
\renewcommand{\topfraction}{0.95}
\renewcommand{\bottomfraction}{0.95}
\renewcommand{\textfraction}{0.05}
\renewcommand{\dbltopfraction}{0.95}
\renewcommand{\floatpagefraction}{0.75}
\renewcommand{\dblfloatpagefraction}{0.75}
\setcounter{topnumber}{4}
\setcounter{bottomnumber}{4}
\setcounter{totalnumber}{8}
\setcounter{dbltopnumber}{4}
\section{Experimental Setup Details}
\label{app:setup-details}

\paragraph{Prompt templates.}
The prompts used for each dataset are shown below.

\begin{tcolorbox}[colback=gray!4,colframe=gray!50,boxrule=0.4pt,arc=1pt,left=5pt,right=5pt,top=4pt,bottom=4pt]
\small\raggedright
\textbf{GSM8K / MATH}\par
{\ttfamily\{question\}}\par\medskip
\textbf{HumanEval}\par
{\ttfamily Complete the following Python function. Return only the completed function.\par\smallskip
\{function prompt\}}\par\medskip
\textbf{FOLIO}\par
{\ttfamily Given the premises, determine whether the conclusion is True, False, or Unknown. Answer with exactly one word: True, False, or Unknown.\par\smallskip
Premises:\par
\{premises\}\par\smallskip
Conclusion: \{conclusion\}}
\end{tcolorbox}

\paragraph{Aggregation and uncertainty.}
Forced-opening means average repetitions within questions, then questions. Natural conditional means pool trajectories by observed opening. For the first-token effect figures, 95\% confidence intervals use 10,000 bootstrap resamples of whole questions, retaining all their trajectories. Experiment-specific aggregation and bootstrap settings are given in the corresponding appendices.

\paragraph{Computing infrastructure.}
We conducted our experiments on NVIDIA RTX A6000 GPUs with 48\,GB of memory and NVIDIA A800 GPUs with 80\,GB of memory.
\section{Additional First-Token Effects}
\label{app:baselines}

\begin{figure*}[!tbp]
\centering
\includegraphics[width=\textwidth]{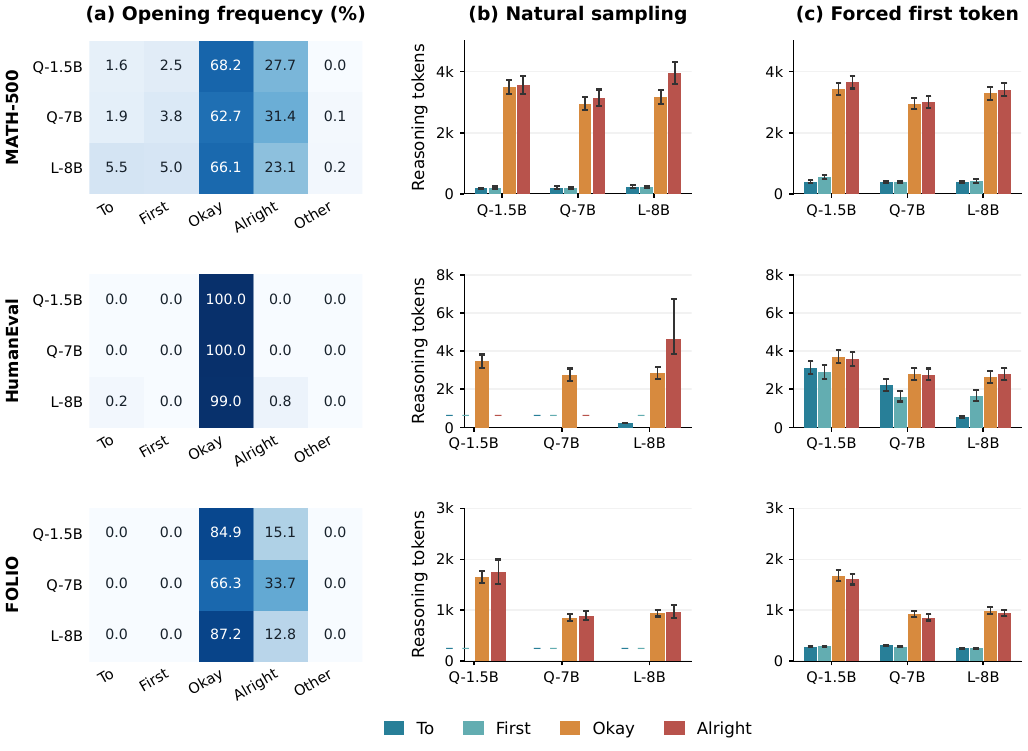}
\caption{First-token effects on MATH-500, HumanEval, and FOLIO (rows). Columns show (a) natural opening frequencies in percent, (b) mean reasoning length by natural opening, and (c) mean reasoning length under forced openings. Length panels share a vertical scale within each dataset. Error bars show 95\% question-bootstrap intervals; dashes indicate openings with no observed natural samples. Other includes all remaining openings. Q and L denote Qwen and Llama.}
\label{fig:baseline-cross-dataset}
\end{figure*}

Figure~\ref{fig:baseline-cross-dataset} extends Figure~\ref{fig:preliminary} to MATH-500, HumanEval, and FOLIO using the reference trajectories described in Appendix~\ref{app:setup-details}. 

The four selected openings account for 92.1--96.8\% of natural trajectories on GSM8K, 99.8--100\% on MATH-500, and 100\% on HumanEval and FOLIO. Although the first-token distribution under natural sampling varies across datasets, forced-opening generation reveals similar reasoning-length contrasts across all four datasets. These results suggest that the first-token effect on reasoning length occurs across diverse tasks.

\section{Additional Prefix-Preserving Intervention Results}
\label{app:prefix}
We supplement Section~\ref{sec:prefix} with length results on MATH-500, HumanEval, and FOLIO, together with accuracy evaluations across all four datasets.

\subsection{Additional Length Results}
Long-to-short comparisons retain questions for which both source trajectories exceed 200 tokens; short-to-long comparisons retain completed sources with at least $K$ tokens. Both directions compare mean remaining lengths under a common horizon of $8{,}192-K$ tokens, including unfinished continuations. We average repetitions within sources, eligible sources within questions, and then questions, using identical weights for the references. In the accompanying tables, \emph{questions} counts distinct problems and \emph{sources} counts eligible original trajectories before continuation resampling, with up to two source first tokens per question.

\begin{figure*}[!tbp]
    \centering
    \includegraphics[width=\textwidth]{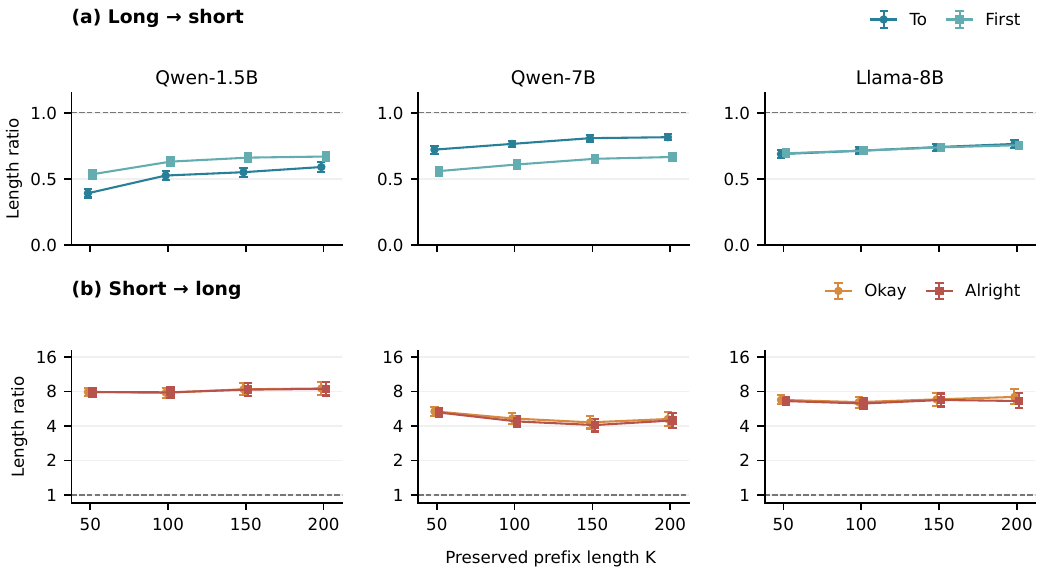}
    \caption{Prefix-preserving first-token replacements on MATH-500. Top: long-to-short replacements across three models. Bottom: short-to-long replacements across the same models. Values are ratios of mean remaining reasoning lengths to matched unchanged-prefix references; error bars show 95\% question-bootstrap intervals.}
    \label{fig:prefix-math}
\end{figure*}

Figure~\ref{fig:prefix-math} reports the MATH-500 results. For long-to-short comparisons, all 164 HumanEval problems are retained, alongside 499, 489, and 488 FOLIO questions for Qwen-1.5B, Qwen-7B, and Llama-8B, respectively. Short-to-long source coverage varies with $K$.

\begin{figure*}[!tbp]
    \centering
    \includegraphics[width=\textwidth]{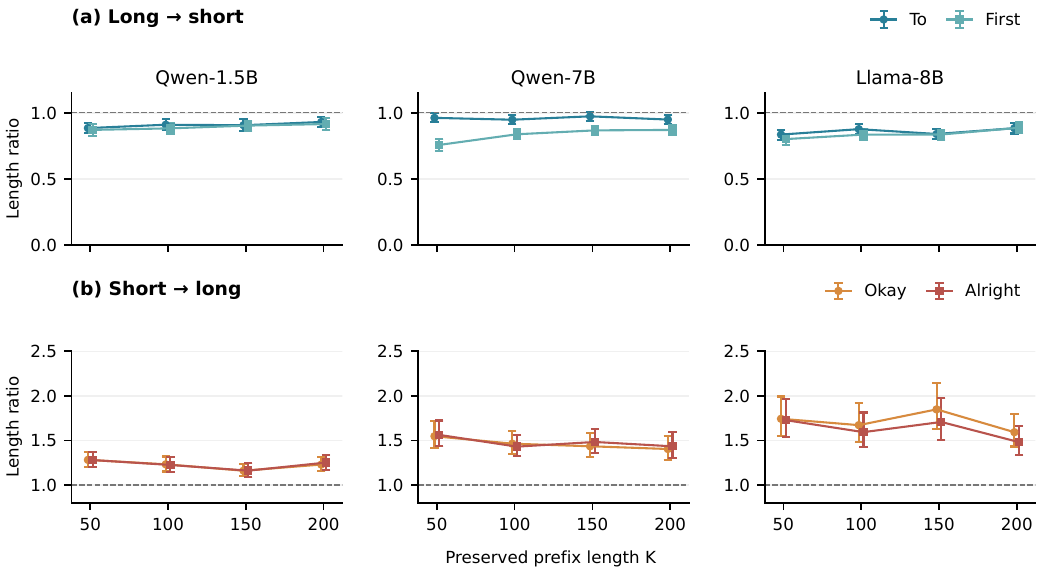}
    \caption{Prefix-preserving first-token replacement on HumanEval. Ratios compare mean edited continuation lengths with the three-continuation unchanged-prefix references. Error bars show 95\% question-bootstrap intervals; dashed lines mark a ratio of one.}
    \label{fig:transfer-prefix-humaneval}
\end{figure*}

\begin{figure*}[!tbp]
    \centering
    \includegraphics[width=\textwidth]{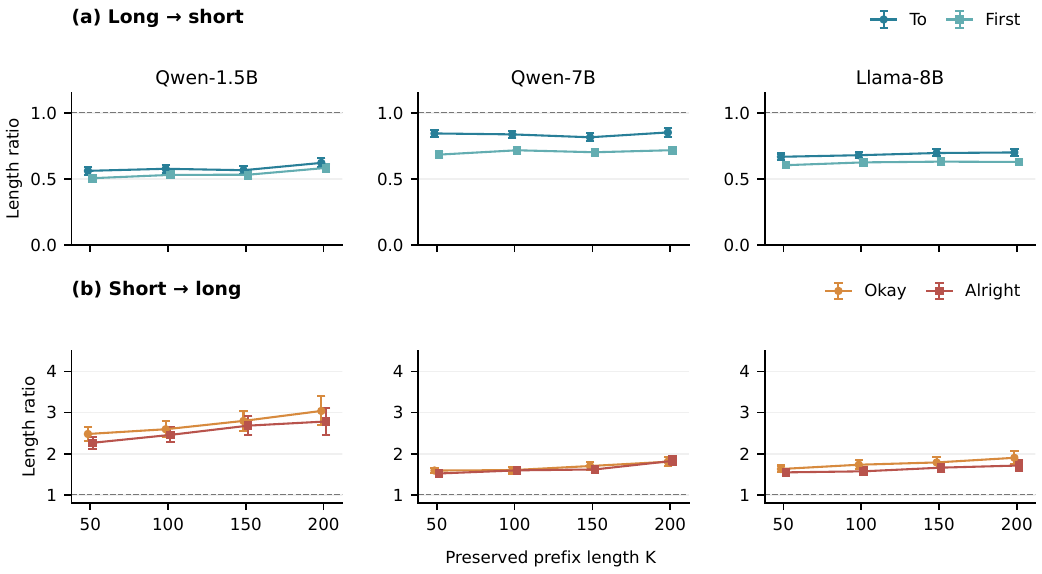}
    \caption{Prefix-preserving first-token replacement on FOLIO, using the same comparison as Figure~\ref{fig:transfer-prefix-humaneval}.}
    \label{fig:transfer-prefix-folio}
\end{figure*}

\paragraph{Length effects beyond mathematical tasks.}
Figures~\ref{fig:transfer-prefix-humaneval} and~\ref{fig:transfer-prefix-folio} show bidirectional mean length changes, with weaker relative shortening on HumanEval than on FOLIO. At $K=200$, HumanEval long-to-short ratios range from 0.87 to 0.95, with all six 95\% intervals below one, while short-to-long ratios range from 1.23 to 1.59. FOLIO shows larger relative changes, with corresponding ratios of 0.58--0.85 and 1.71--3.04. These results extend the persistence observation to coding and logical reasoning, while showing that its magnitude depends on the task and replacement.

\subsection{Accuracy Evaluation}
Accuracy is evaluated on the same eligible sources as the length comparisons, with each sampled continuation scored separately. Unfinished reasoning and missing or unparseable answers count as incorrect.

\paragraph{GSM8K and MATH-500.}
Tables~\ref{tab:prefix-results-gsm8k} and~\ref{tab:prefix-results-math} report accuracy on the mathematical datasets. For GSM8K, we extract answers after \texttt{</think>}, prioritizing the last complete \texttt{\textbackslash boxed} expression and otherwise the last number, with fractions and unit formatting normalized. For MATH-500, extraction prioritizes boxed answers and otherwise uses an explicit final-answer section when available; Markdown emphasis, ordinal suffixes, and degree notation are normalized.

\paragraph{HumanEval and FOLIO.}
Tables~\ref{tab:transfer-accuracy-humaneval} and~\ref{tab:transfer-accuracy-folio} report accuracy on the additional datasets. HumanEval programs are extracted from final answers, retaining generated imports and helper functions as well as imports supplied in the task prompt. Programs are evaluated against all supplied task tests in an isolated environment; test failures or execution exceeding 3 seconds of CPU time or 8 seconds of wall-clock time count as incorrect. FOLIO predictions are extracted from final-answer labels, prioritizing explicit answers over labels mentioned within explanations.

\FloatBarrier
\begin{table*}[!tp]
\centering\small
\setlength{\tabcolsep}{5pt}
\begin{tabular}{llrrrrrrr}
\toprule
\rowcolor{black!10}\multicolumn{9}{c}{Long-to-short (Okay / Alright $\rightarrow$ To / First)} \\
Model & $K$ & Questions / sources & \multicolumn{2}{c}{Reference} & \multicolumn{2}{c}{\texttt{To}} & \multicolumn{2}{c}{\texttt{First}} \\
\cmidrule(lr){4-5}\cmidrule(lr){6-7}\cmidrule(lr){8-9}
 & & & Length & Acc. & Length & Acc. & Length & Acc. \\
\midrule
Qwen-1.5B & 50 & 500 / 1000 & 1,608.6 & 88.5 & 274.5 & 86.3 & 343.2 & 87.6 \\
 & 100 & 500 / 1000 & 1,565.4 & 89.2 & 358.5 & 86.7 & 432.5 & 87.8 \\
 & 150 & 500 / 1000 & 1,472.6 & 89.9 & 404.5 & 87.1 & 489.8 & 87.8 \\
 & 200 & 500 / 1000 & 1,461.2 & 88.6 & 419.5 & 86.3 & 524.7 & 86.9 \\
\addlinespace[3pt]
Qwen-7B & 50 & 500 / 1000 & 1,138.5 & 95.0 & 414.2 & 94.8 & 306.7 & 94.4 \\
 & 100 & 500 / 1000 & 1,096.9 & 94.9 & 490.7 & 94.1 & 336.6 & 94.5 \\
 & 150 & 500 / 1000 & 1,059.4 & 94.9 & 507.5 & 93.5 & 349.4 & 94.3 \\
 & 200 & 500 / 1000 & 1,019.2 & 94.8 & 482.6 & 93.7 & 347.2 & 94.5 \\
\addlinespace[3pt]
Llama-8B & 50 & 500 / 1000 & 1,503.8 & 88.5 & 579.8 & 84.9 & 567.7 & 85.1 \\
 & 100 & 500 / 1000 & 1,442.1 & 88.7 & 681.8 & 85.8 & 680.3 & 85.8 \\
 & 150 & 500 / 1000 & 1,430.8 & 88.0 & 686.5 & 86.4 & 682.6 & 86.2 \\
 & 200 & 500 / 1000 & 1,348.0 & 89.0 & 683.3 & 86.6 & 665.6 & 85.9 \\
\midrule
\rowcolor{black!10}\multicolumn{9}{c}{Short-to-long (To / First $\rightarrow$ Okay / Alright)} \\
Model & $K$ & Questions / sources & \multicolumn{2}{c}{Reference} & \multicolumn{2}{c}{\texttt{Okay}} & \multicolumn{2}{c}{\texttt{Alright}} \\
\cmidrule(lr){4-5}\cmidrule(lr){6-7}\cmidrule(lr){8-9}
 & & & Length & Acc. & Length & Acc. & Length & Acc. \\
\midrule
Qwen-1.5B & 50 & 500 / 999 & 117.7 & 82.8 & 744.4 & 85.4 & 734.7 & 86.2 \\
 & 100 & 492 / 932 & 70.4 & 83.0 & 441.7 & 85.0 & 421.7 & 84.8 \\
 & 150 & 352 / 562 & 47.6 & 80.2 & 374.4 & 81.8 & 356.0 & 82.1 \\
 & 200 & 165 / 223 & 48.1 & 74.1 & 510.4 & 75.4 & 495.9 & 77.4 \\
\addlinespace[3pt]
Qwen-7B & 50 & 500 / 999 & 120.2 & 92.0 & 181.0 & 92.9 & 180.4 & 93.4 \\
 & 100 & 496 / 929 & 74.9 & 92.2 & 102.4 & 92.9 & 101.8 & 92.9 \\
 & 150 & 372 / 597 & 49.5 & 89.6 & 79.1 & 90.6 & 68.2 & 90.3 \\
 & 200 & 170 / 240 & 43.2 & 82.8 & 75.4 & 83.0 & 62.3 & 82.4 \\
\addlinespace[3pt]
Llama-8B & 50 & 500 / 1000 & 108.4 & 71.3 & 306.9 & 76.2 & 271.8 & 75.5 \\
 & 100 & 483 / 894 & 64.3 & 70.9 & 192.9 & 74.4 & 150.9 & 73.1 \\
 & 150 & 328 / 494 & 46.8 & 64.7 & 170.8 & 67.9 & 151.0 & 67.5 \\
 & 200 & 138 / 181 & 47.5 & 54.5 & 235.2 & 56.3 & 193.9 & 55.4 \\
\bottomrule
\end{tabular}
\caption{Prefix-preserving interventions on GSM8K. Length is mean remaining reasoning tokens; Acc. is answer accuracy (\%). Reference averages the observed source outcome and two unchanged-prefix continuations; each replacement averages three continuations. All quantities use matched eligible sources and equal question weights.}
\label{tab:prefix-results-gsm8k}
\label{tab:prefix-accuracy-all}
\label{tab:revflip}
\end{table*}

\begin{table*}[!tp]
\centering\small
\setlength{\tabcolsep}{5pt}
\begin{tabular}{llrrrrrrr}
\toprule
\rowcolor{black!10}\multicolumn{9}{c}{Long-to-short (Okay / Alright $\rightarrow$ To / First)} \\
Model & $K$ & Questions / sources & \multicolumn{2}{c}{Reference} & \multicolumn{2}{c}{\texttt{To}} & \multicolumn{2}{c}{\texttt{First}} \\
\cmidrule(lr){4-5}\cmidrule(lr){6-7}\cmidrule(lr){8-9}
 & & & Length & Acc. & Length & Acc. & Length & Acc. \\
\midrule
Qwen-1.5B & 50 & 500 / 1000 & 3,503.2 & 78.0 & 1,367.5 & 73.5 & 1,872.0 & 76.0 \\
 & 100 & 500 / 1000 & 3,452.4 & 77.8 & 1,810.6 & 75.4 & 2,174.3 & 77.0 \\
 & 150 & 500 / 1000 & 3,406.4 & 78.7 & 1,872.7 & 75.1 & 2,248.6 & 77.1 \\
 & 200 & 500 / 1000 & 3,342.6 & 78.5 & 1,968.3 & 76.4 & 2,234.3 & 77.3 \\
\addlinespace[3pt]
Qwen-7B & 50 & 500 / 1000 & 2,959.6 & 88.3 & 2,132.8 & 88.4 & 1,652.1 & 87.1 \\
 & 100 & 500 / 1000 & 2,914.7 & 88.3 & 2,227.3 & 88.2 & 1,772.6 & 87.3 \\
 & 150 & 500 / 1000 & 2,837.4 & 88.5 & 2,290.6 & 87.3 & 1,847.4 & 88.1 \\
 & 200 & 500 / 1000 & 2,796.3 & 88.6 & 2,278.1 & 88.2 & 1,859.2 & 87.8 \\
\addlinespace[3pt]
Llama-8B & 50 & 500 / 1000 & 3,253.0 & 81.7 & 2,235.6 & 79.3 & 2,253.3 & 79.7 \\
 & 100 & 500 / 1000 & 3,233.4 & 81.3 & 2,301.6 & 81.0 & 2,305.9 & 80.5 \\
 & 150 & 500 / 1000 & 3,179.5 & 80.9 & 2,351.7 & 80.7 & 2,346.3 & 80.5 \\
 & 200 & 500 / 1000 & 3,101.2 & 81.4 & 2,368.0 & 80.8 & 2,339.2 & 81.1 \\
\midrule
\rowcolor{black!10}\multicolumn{9}{c}{Short-to-long (To / First $\rightarrow$ Okay / Alright)} \\
Model & $K$ & Questions / sources & \multicolumn{2}{c}{Reference} & \multicolumn{2}{c}{\texttt{Okay}} & \multicolumn{2}{c}{\texttt{Alright}} \\
\cmidrule(lr){4-5}\cmidrule(lr){6-7}\cmidrule(lr){8-9}
 & & & Length & Acc. & Length & Acc. & Length & Acc. \\
\midrule
Qwen-1.5B & 50 & 500 / 989 & 354.0 & 66.5 & 2,805.8 & 77.6 & 2,794.6 & 77.4 \\
 & 100 & 500 / 973 & 312.0 & 65.9 & 2,429.9 & 76.5 & 2,451.8 & 76.7 \\
 & 150 & 472 / 870 & 275.6 & 66.0 & 2,302.0 & 75.4 & 2,288.2 & 76.0 \\
 & 200 & 420 / 736 & 266.2 & 62.8 & 2,260.0 & 70.9 & 2,239.5 & 72.1 \\
\addlinespace[3pt]
Qwen-7B & 50 & 500 / 995 & 333.1 & 75.8 & 1,785.8 & 86.4 & 1,743.6 & 86.5 \\
 & 100 & 496 / 978 & 282.8 & 75.1 & 1,314.8 & 84.2 & 1,237.5 & 84.4 \\
 & 150 & 473 / 884 & 257.1 & 74.3 & 1,104.8 & 83.2 & 1,046.1 & 82.2 \\
 & 200 & 412 / 735 & 240.4 & 71.3 & 1,104.2 & 79.9 & 1,073.5 & 78.4 \\
\addlinespace[3pt]
Llama-8B & 50 & 500 / 994 & 336.8 & 58.6 & 2,272.3 & 78.1 & 2,219.4 & 77.6 \\
 & 100 & 495 / 966 & 296.5 & 58.0 & 1,913.1 & 75.1 & 1,860.4 & 73.5 \\
 & 150 & 462 / 867 & 269.3 & 56.1 & 1,836.6 & 71.0 & 1,814.7 & 70.1 \\
 & 200 & 398 / 707 & 278.1 & 53.4 & 1,991.5 & 66.8 & 1,835.3 & 65.5 \\
\bottomrule
\end{tabular}
\caption{Prefix-preserving interventions on MATH-500. Length is mean remaining reasoning tokens; Acc. is answer accuracy (\%). Reference averages the observed source outcome and two unchanged-prefix continuations; each replacement averages three continuations. All quantities use matched eligible sources and equal question weights.}
\label{tab:prefix-results-math}
\end{table*}

\begin{table*}[!tp]
\centering\small
\setlength{\tabcolsep}{5pt}
\begin{tabular}{llrrrrrrr}
\toprule
\rowcolor{black!10}\multicolumn{9}{c}{Long-to-short (Okay / Alright $\rightarrow$ To / First)} \\
Model & $K$ & Questions / sources & \multicolumn{2}{c}{Reference} & \multicolumn{2}{c}{\texttt{To}} & \multicolumn{2}{c}{\texttt{First}} \\
\cmidrule(lr){4-5}\cmidrule(lr){6-7}\cmidrule(lr){8-9}
 & & & Length & Acc. & Length & Acc. & Length & Acc. \\
\midrule
Qwen-1.5B & 50 & 164 / 328 & 3,554.1 & 64.7 & 3,140.9 & 64.9 & 3,092.6 & 63.9 \\
 & 100 & 164 / 328 & 3,589.9 & 63.7 & 3,263.8 & 63.2 & 3,165.6 & 64.2 \\
 & 150 & 164 / 328 & 3,441.8 & 64.6 & 3,116.0 & 66.5 & 3,106.7 & 66.6 \\
 & 200 & 164 / 328 & 3,426.0 & 65.0 & 3,186.4 & 64.7 & 3,133.2 & 62.7 \\
\addlinespace[3pt]
Qwen-7B & 50 & 164 / 328 & 2,758.8 & 85.9 & 2,651.8 & 83.1 & 2,085.9 & 83.7 \\
 & 100 & 164 / 328 & 2,671.6 & 85.9 & 2,530.3 & 86.2 & 2,236.1 & 84.5 \\
 & 150 & 164 / 328 & 2,613.4 & 86.9 & 2,542.3 & 86.2 & 2,264.2 & 84.2 \\
 & 200 & 164 / 328 & 2,586.6 & 85.1 & 2,452.8 & 86.5 & 2,249.8 & 85.1 \\
\addlinespace[3pt]
Llama-8B & 50 & 164 / 328 & 2,707.7 & 84.5 & 2,260.0 & 81.7 & 2,166.9 & 81.3 \\
 & 100 & 164 / 328 & 2,631.9 & 84.5 & 2,304.7 & 81.9 & 2,196.7 & 81.6 \\
 & 150 & 164 / 328 & 2,675.0 & 82.6 & 2,244.6 & 82.8 & 2,233.9 & 82.7 \\
 & 200 & 164 / 328 & 2,551.4 & 83.5 & 2,252.8 & 82.6 & 2,261.0 & 81.6 \\
\midrule
\rowcolor{black!10}\multicolumn{9}{c}{Short-to-long (To / First $\rightarrow$ Okay / Alright)} \\
Model & $K$ & Questions / sources & \multicolumn{2}{c}{Reference} & \multicolumn{2}{c}{\texttt{Okay}} & \multicolumn{2}{c}{\texttt{Alright}} \\
\cmidrule(lr){4-5}\cmidrule(lr){6-7}\cmidrule(lr){8-9}
 & & & Length & Acc. & Length & Acc. & Length & Acc. \\
\midrule
Qwen-1.5B & 50 & 157 / 293 & 2,601.3 & 67.9 & 3,335.1 & 67.5 & 3,330.8 & 68.9 \\
 & 100 & 157 / 293 & 2,629.0 & 66.7 & 3,240.3 & 68.6 & 3,221.1 & 68.6 \\
 & 150 & 157 / 293 & 2,657.2 & 66.5 & 3,085.1 & 67.5 & 3,086.4 & 67.4 \\
 & 200 & 157 / 290 & 2,456.9 & 68.2 & 3,021.9 & 68.3 & 3,076.1 & 67.2 \\
\addlinespace[3pt]
Qwen-7B & 50 & 161 / 312 & 1,684.1 & 84.4 & 2,605.5 & 85.1 & 2,630.6 & 85.1 \\
 & 100 & 161 / 312 & 1,727.9 & 84.5 & 2,528.6 & 86.0 & 2,471.6 & 86.2 \\
 & 150 & 161 / 308 & 1,686.3 & 84.5 & 2,419.2 & 87.2 & 2,500.5 & 86.4 \\
 & 200 & 157 / 297 & 1,690.3 & 84.4 & 2,373.1 & 85.7 & 2,423.7 & 85.8 \\
\addlinespace[3pt]
Llama-8B & 50 & 160 / 190 & 1,402.2 & 76.1 & 2,442.9 & 85.4 & 2,421.8 & 83.9 \\
 & 100 & 159 / 189 & 1,423.7 & 76.2 & 2,379.0 & 81.8 & 2,267.4 & 81.0 \\
 & 150 & 157 / 184 & 1,262.1 & 76.2 & 2,331.6 & 82.8 & 2,153.8 & 82.1 \\
 & 200 & 148 / 171 & 1,431.3 & 76.5 & 2,278.7 & 83.3 & 2,119.2 & 81.1 \\
\bottomrule
\end{tabular}
\caption{Prefix-preserving interventions on HumanEval. Length is mean remaining reasoning tokens; Acc. is answer accuracy (\%). Reference averages the observed source outcome and two unchanged-prefix continuations; each replacement averages three continuations. All quantities use matched eligible sources and equal question weights.}
\label{tab:prefix-results-humaneval}
\label{tab:transfer-prefix-humaneval}
\label{tab:transfer-accuracy-humaneval}
\end{table*}

\begin{table*}[!tp]
\centering\small
\setlength{\tabcolsep}{5pt}
\begin{tabular}{llrrrrrrr}
\toprule
\rowcolor{black!10}\multicolumn{9}{c}{Long-to-short (Okay / Alright $\rightarrow$ To / First)} \\
Model & $K$ & Questions / sources & \multicolumn{2}{c}{Reference} & \multicolumn{2}{c}{\texttt{To}} & \multicolumn{2}{c}{\texttt{First}} \\
\cmidrule(lr){4-5}\cmidrule(lr){6-7}\cmidrule(lr){8-9}
 & & & Length & Acc. & Length & Acc. & Length & Acc. \\
\midrule
Qwen-1.5B & 50 & 499 / 998 & 1,580.1 & 44.6 & 884.7 & 42.1 & 796.9 & 41.5 \\
 & 100 & 499 / 998 & 1,531.2 & 44.0 & 881.3 & 42.4 & 810.5 & 41.3 \\
 & 150 & 499 / 998 & 1,515.9 & 43.4 & 856.0 & 44.1 & 804.4 & 42.2 \\
 & 200 & 499 / 998 & 1,402.6 & 43.6 & 869.9 & 42.2 & 816.5 & 42.6 \\
\addlinespace[3pt]
Qwen-7B & 50 & 489 / 978 & 835.2 & 51.2 & 704.2 & 49.0 & 570.4 & 48.1 \\
 & 100 & 489 / 978 & 785.8 & 50.4 & 657.0 & 50.1 & 562.4 & 48.5 \\
 & 150 & 489 / 978 & 748.0 & 49.9 & 609.0 & 49.8 & 524.2 & 48.6 \\
 & 200 & 489 / 978 & 676.8 & 50.4 & 575.9 & 49.3 & 484.9 & 48.6 \\
\addlinespace[3pt]
Llama-8B & 50 & 488 / 976 & 934.2 & 49.1 & 622.6 & 47.9 & 563.1 & 46.8 \\
 & 100 & 488 / 976 & 880.9 & 49.6 & 597.5 & 47.8 & 549.8 & 47.0 \\
 & 150 & 488 / 976 & 820.8 & 49.4 & 570.4 & 47.6 & 516.8 & 47.7 \\
 & 200 & 488 / 976 & 775.0 & 49.1 & 541.9 & 48.3 & 486.1 & 47.2 \\
\midrule
\rowcolor{black!10}\multicolumn{9}{c}{Short-to-long (To / First $\rightarrow$ Okay / Alright)} \\
Model & $K$ & Questions / sources & \multicolumn{2}{c}{Reference} & \multicolumn{2}{c}{\texttt{Okay}} & \multicolumn{2}{c}{\texttt{Alright}} \\
\cmidrule(lr){4-5}\cmidrule(lr){6-7}\cmidrule(lr){8-9}
 & & & Length & Acc. & Length & Acc. & Length & Acc. \\
\midrule
Qwen-1.5B & 50 & 500 / 999 & 234.6 & 40.3 & 581.9 & 42.6 & 531.6 & 41.7 \\
 & 100 & 500 / 995 & 188.6 & 40.3 & 489.6 & 42.4 & 463.6 & 42.1 \\
 & 150 & 491 / 920 & 144.5 & 40.6 & 404.6 & 41.6 & 387.9 & 41.6 \\
 & 200 & 441 / 753 & 128.7 & 38.3 & 391.5 & 40.2 & 358.5 & 40.3 \\
\addlinespace[3pt]
Qwen-7B & 50 & 500 / 1000 & 245.6 & 46.6 & 390.5 & 49.4 & 372.8 & 48.8 \\
 & 100 & 500 / 990 & 193.8 & 46.9 & 310.0 & 50.1 & 308.9 & 48.8 \\
 & 150 & 485 / 913 & 155.9 & 44.5 & 265.0 & 47.1 & 252.0 & 46.9 \\
 & 200 & 434 / 761 & 124.0 & 42.3 & 223.8 & 45.5 & 226.3 & 44.6 \\
\addlinespace[3pt]
Llama-8B & 50 & 500 / 1000 & 201.3 & 43.5 & 327.6 & 47.6 & 311.1 & 46.5 \\
 & 100 & 500 / 991 & 153.4 & 44.3 & 265.6 & 46.9 & 241.0 & 46.2 \\
 & 150 & 486 / 914 & 110.8 & 42.5 & 198.0 & 44.2 & 184.0 & 44.9 \\
 & 200 & 415 / 688 & 87.2 & 39.1 & 165.6 & 40.7 & 149.3 & 39.3 \\
\bottomrule
\end{tabular}
\caption{Prefix-preserving interventions on FOLIO. Length is mean remaining reasoning tokens; Acc. is answer accuracy (\%). Reference averages the observed source outcome and two unchanged-prefix continuations; each replacement averages three continuations. All quantities use matched eligible sources and equal question weights.}
\label{tab:prefix-results-folio}
\label{tab:transfer-prefix-folio}
\label{tab:transfer-accuracy-folio}
\end{table*}


\subsection{Illustrative Prefix-Preserving Interventions}
\label{app:prefix-cases}
Figures~\ref{fig:prefix-case-prefix}--\ref{fig:transfer-case-folio-reverse} show Qwen-7B cases in both directions on GSM8K, HumanEval, and FOLIO at $K=200$. QIDs are zero-based dataset row indices and coincide with numeric task IDs for HumanEval.

\FloatBarrier
\begin{figure*}[!p]

\begin{tcolorbox}[colback=black!1,colframe=black!25,boxrule=0.5pt,arc=1pt,left=7pt,right=7pt,top=6pt,bottom=6pt]
\small
\textbf{Long-to-short. Qwen-7B $\cdot$ GSM8K QID 6656 $\cdot$ $K=200$}\par\medskip
\textbf{Question.} The pet store can buy a goldfish for \$.25 and sell it for \$.75. The owner plans to use the profits from goldfish sales to buy a new tank, which costs \$100. After one week, he is 45\% short of the price. How many goldfish did he sell that week?\par\medskip
\textbf{First-token replacement.}
\fcolorbox{black!20}{gray!10}{\strut\textcolor{caseSource}{\textbf{\texttt{Okay}}} $\rightarrow$ \textcolor{caseEdited}{\textbf{\texttt{First}}}}\par\smallskip
\textbf{Shared prefix $r_{2:200}$ (199 tokens; excerpts shown).}\par , so I need to figure out how many goldfish the pet store owner sold in one week. Let me read the problem again and make sure I understand all the details.\textit{[\ldots]} . So, he hasn't reached \$100 yet, but he's missing 45\% of it. I need to figure out how\par\medskip
\begin{minipage}[t]{0.48\linewidth}\vspace{0pt}
\begin{tcolorbox}[colback=caseSource!6,colframe=caseSource!30,boxrule=0.3pt,arc=1pt,boxsep=0pt,left=4pt,right=4pt,top=4pt,bottom=4pt,before skip=0pt,after skip=0pt]
\textcolor{caseSource}{\textbf{Original continuation}}\par
\textcolor{caseSource}{\textbf{659 tokens}}\textit{; excerpts}\par\smallskip
 much he has so far and then determine how many goldfish he sold to make that amount.\textit{[\ldots]} Wait, so does that mean he sold 110 goldfish? \textcolor{caseSource}{\textbf{Let me double-check my steps.}}\textit{[\ldots]} \textcolor{caseSource}{\textbf{Wait, let me think again.}} Is 55\% of \$100 actually \$55?\textit{[\ldots]} \textcolor{caseSource}{\textbf{But let me make sure that 110 goldfish is the right number.}} Let's compute it again: \textit{[\ldots]}
\end{tcolorbox}
\end{minipage}\hfill
\begin{minipage}[t]{0.48\linewidth}\vspace{0pt}
\begin{tcolorbox}[colback=caseEdited!6,colframe=caseEdited!30,boxrule=0.3pt,arc=1pt,boxsep=0pt,left=4pt,right=4pt,top=4pt,bottom=4pt,before skip=0pt,after skip=0pt]
\textcolor{caseEdited}{\textbf{Edited continuation}}\par
\textcolor{caseEdited}{\textbf{224 tokens}}\textit{; excerpts}\par\smallskip
 much he has saved so far and then determine how many goldfish he sold to make that amount.\textit{[\ldots]} Number of goldfish = Amount saved / Profit per goldfish\newline = \$55 / \$0.50\newline = 110\par\smallskip \textcolor{caseEdited}{\textbf{Therefore, the owner sold 110 goldfish in one week.}}
\end{tcolorbox}
\end{minipage}\par\medskip
\textbf{Final answer.} Both return 110 (correct).\par
\textbf{All three edited samples.} 224, 218, 309 remaining reasoning tokens.
\end{tcolorbox}
\caption{Long-to-short prefix-preserving intervention. Tokens $r_2,\ldots,r_{200}$ are identical in the two conditions. Bracketed ellipses mark omitted text; lengths count the entire reasoning continuation after position 200, excluding the final answer. The source revisits the calculation several times, whereas the edited continuation reaches the same answer with less repeated checking.}
\label{fig:prefix-case-prefix}

\par\vspace{10pt}

\begin{tcolorbox}[colback=black!1,colframe=black!25,boxrule=0.5pt,arc=1pt,left=7pt,right=7pt,top=6pt,bottom=6pt]
\small
\textbf{Short-to-long. Qwen-7B $\cdot$ GSM8K QID 2549 $\cdot$ $K=200$}\par\medskip
\textbf{Question.} A blacksmith has 400kg of iron to make horseshoes for neighbouring farms and stables. There are 2 farms nearby, each of which has 2 horses. There are 2 stables nearby which all have 5 horses each. While the blacksmith is crafting, a riding school reaches out and asks for as many horseshoes as possible for their own horses. The blacksmith promises to give them however many horseshoes he has left when he has finished his orders for the farms and stables. If each horseshoe needs 2kg of iron, how many horses will get new horseshoes at the riding school?\par\medskip
\textbf{First-token replacement.}
\fcolorbox{black!20}{gray!10}{\strut\textcolor{caseSource}{\textbf{\texttt{First}}} $\rightarrow$ \textcolor{caseEdited}{\textbf{\texttt{Okay}}}}\par\smallskip
\textbf{Shared prefix $r_{2:200}$ (199 tokens; excerpts shown).}\par , determine the total amount of iron available, which is 400 kg.\textit{[\ldots]}  the total iron:\newline 400 kg - 112 kg = 288 kg remaining.\par\smallskip Finally, determine how many horses can be\par\medskip
\begin{minipage}[t]{0.48\linewidth}\vspace{0pt}
\begin{tcolorbox}[colback=caseSource!6,colframe=caseSource!30,boxrule=0.3pt,arc=1pt,boxsep=0pt,left=4pt,right=4pt,top=4pt,bottom=4pt,before skip=0pt,after skip=0pt]
\textcolor{caseSource}{\textbf{Original continuation}}\par
\textcolor{caseSource}{\textbf{68 tokens}}\textit{; complete reasoning suffix}\par\smallskip
provided horseshoes with the remaining iron:\newline 288 kg $\div$ 2 kg/horseshoe = 144 horseshoes.\newline 144 horseshoes $\div$ 4 horseshoes/horse = 36 horses.\par\smallskip Therefore, the riding school will receive horseshoes for 36 horses.
\end{tcolorbox}
\end{minipage}\hfill
\begin{minipage}[t]{0.48\linewidth}\vspace{0pt}
\begin{tcolorbox}[colback=caseEdited!6,colframe=caseEdited!30,boxrule=0.3pt,arc=1pt,boxsep=0pt,left=4pt,right=4pt,top=4pt,bottom=4pt,before skip=0pt,after skip=0pt]
\textcolor{caseEdited}{\textbf{Edited continuation}}\par
\textcolor{caseEdited}{\textbf{91 tokens}}\textit{; complete reasoning suffix}\par\smallskip
provided with horseshoes from the remaining iron:\newline 288 kg $\div$ 2 kg/horseshoe = 144 horseshoes.\par\smallskip \textcolor{caseEdited}{\textbf{Since each horse needs 4 horseshoes, divide the number of horseshoes by 4:}}\newline 144 horseshoes $\div$ 4 horseshoes/horse = 36 horses.\par\smallskip Therefore, the riding school will receive new horseshoes for 36 horses.
\end{tcolorbox}
\end{minipage}\par\medskip
\textbf{Final answer.} Both return 36 (correct).\par
\textbf{All three edited samples.} 91, 77, 96 remaining reasoning tokens.
\end{tcolorbox}
\caption{Short-to-long prefix-preserving intervention. Tokens $r_2,\ldots,r_{200}$ are identical in the two conditions. Bracketed ellipses mark omitted text; lengths count the entire reasoning continuation after position 200, excluding the final answer. The edited continuation adds an explicit explanation of the division by four while retaining the same calculation and answer.}
\label{fig:prefix-case-revflip}

\end{figure*}

\begin{figure*}[!p]

\begin{tcolorbox}[colback=black!1,colframe=black!25,boxrule=0.5pt,arc=1pt,left=7pt,right=7pt,top=6pt,bottom=6pt]
\small
\textbf{HumanEval long-to-short. Qwen-7B $\cdot$ QID 147 $\cdot$ $K=200$}\par\smallskip
\textbf{Task summary.} Implement \texttt{get\_max\_triples(n)}. Set $a_i=i^2-i+1$ for $1\leq i\leq n$, and count triples $i<j<k$ whose values sum to a multiple of three.\par\smallskip
\textbf{First-token replacement.} \textcolor{caseSource}{\textbf{\texttt{Okay}}} $\rightarrow$ \textcolor{caseEdited}{\textbf{\texttt{First}}}\par\smallskip
\textbf{Shared prefix $r_{2:200}$ (199 tokens; excerpts shown).}\par , I need to solve this problem where I have to find the number of triples (a[i], a[j], a[k]) in an array a of length n, such that i < j < k and the sum of a[i], a[j], a[k] is a multiple of 3. The array a is defined such that a[i] = i\textsuperscript{2} - i + 1 for each i from 1 to n. \textit{[\ldots]} 7\newline i=4:16-4+1=13\newline i=5:25-5+1=21\par\medskip
\begin{minipage}[t]{0.48\linewidth}\vspace{0pt}
\begin{tcolorbox}[colback=caseSource!6,colframe=caseSource!30,boxrule=0.3pt,arc=1pt,boxsep=0pt,left=4pt,right=4pt,top=4pt,bottom=4pt,before skip=0pt,after skip=0pt]
\textcolor{caseSource}{\textbf{Original continuation}}\par\textcolor{caseSource}{\textbf{5535 tokens}}\textit{; excerpts}\par\smallskip Now, I need to find triples where the sum is divisible by 3. So, the sum mod 3 should be zero. \textit{[\ldots]} \textcolor{caseSource}{\textbf{But wait, for n=5, the code returns 1, which is correct.}} \textit{[\ldots]} But wait, in the code, the initial condition is if n <3: return 0.\par\smallskip But wait, for n=3, the function returns 0, which is correct. \textit{[\ldots]}
\end{tcolorbox}
\end{minipage}\hfill
\begin{minipage}[t]{0.48\linewidth}\vspace{0pt}
\begin{tcolorbox}[colback=caseEdited!6,colframe=caseEdited!30,boxrule=0.3pt,arc=1pt,boxsep=0pt,left=4pt,right=4pt,top=4pt,bottom=4pt,before skip=0pt,after skip=0pt]
\textcolor{caseEdited}{\textbf{Edited continuation}}\par\textcolor{caseEdited}{\textbf{3561 tokens}}\textit{; excerpts}\par\smallskip Now, the problem requires that the sum of three elements is divisible by 3. \textit{[\ldots]} \textcolor{caseEdited}{\textbf{If i mod3 is 0 or 1, a[i] mod3 is 1.}}\newline If i mod3 is 2, a[i] mod3 is 0. \textit{[\ldots]} Testing with the sample input, it returns 1, which is correct.\par\smallskip Another test: n=0? But n is positive, so no need.\par\smallskip Another test: n=1 $\rightarrow$ returns 0.\par\smallskip n=6 $\rightarrow$ returns 4. \textit{[\ldots]}
\end{tcolorbox}
\end{minipage}\par\medskip
\textbf{Answer evaluation.} The reference and all three edited programs pass the supplied task tests.\par
\textbf{All three edited samples.} 3561, 2961, 5289 remaining reasoning tokens.
\end{tcolorbox}
\caption{HumanEval long-to-short prefix replacement. Bracketed ellipses indicate omissions; lengths count the full reasoning suffix after position 200. Both continuations develop a modulo-three solution and check examples, with different amounts of elaboration.}
\label{fig:transfer-case-humaneval-forward}

\par\vspace{10pt}

\begin{tcolorbox}[colback=black!1,colframe=black!25,boxrule=0.5pt,arc=1pt,left=7pt,right=7pt,top=6pt,bottom=6pt]
\small
\textbf{HumanEval short-to-long. Qwen-7B $\cdot$ QID 97 $\cdot$ $K=200$}\par\smallskip
\textbf{Task summary.} Implement \texttt{multiply(a, b)}, returning the product of the unit digits of two integers, including negative inputs.\par\smallskip
\textbf{First-token replacement.} \textcolor{caseSource}{\textbf{\texttt{First}}} $\rightarrow$ \textcolor{caseEdited}{\textbf{\texttt{Okay}}}\par\smallskip
\textbf{Shared prefix $r_{2:200}$ (199 tokens; excerpts shown).}\par , I need to write a Python function called multiply that takes two integers, a and b, and returns the product of their unit digits. \textit{[\ldots]} 5, which is correct. So taking the modulus works for both positive and negative numbers.\par\smallskip I should also handle cases where the result is negative. However\par\medskip
\begin{minipage}[t]{0.48\linewidth}\vspace{0pt}
\begin{tcolorbox}[colback=caseSource!6,colframe=caseSource!30,boxrule=0.3pt,arc=1pt,boxsep=0pt,left=4pt,right=4pt,top=4pt,bottom=4pt,before skip=0pt,after skip=0pt]
\textcolor{caseSource}{\textbf{Original continuation}}\par\textcolor{caseSource}{\textbf{241 tokens}}\textit{; excerpts}\par\smallskip So I don't need to worry about negative results here; the product is the multiplication of the absolute unit digits. \textit{[\ldots]} Let me test this logic with the examples: \textit{[\ldots]} Example 4: multiply(14, -15)\newline Unit digits are 4 and 5. 4 * 5 = 20. Correct.
\end{tcolorbox}
\end{minipage}\hfill
\begin{minipage}[t]{0.48\linewidth}\vspace{0pt}
\begin{tcolorbox}[colback=caseEdited!6,colframe=caseEdited!30,boxrule=0.3pt,arc=1pt,boxsep=0pt,left=4pt,right=4pt,top=4pt,bottom=4pt,before skip=0pt,after skip=0pt]
\textcolor{caseEdited}{\textbf{Edited continuation}}\par\textcolor{caseEdited}{\textbf{418 tokens}}\textit{; excerpts}\par\smallskip So the steps are: get the last digit of a, last digit of b, multiply them, return the result. \textit{[\ldots]} \textcolor{caseEdited}{\textbf{Wait, but wait: what if a is negative?}} Let me see. For example, a is -14, then a\%10 is 6, right? \textit{[\ldots]} Yes, so that's correct. So the modulus works regardless of the sign.
\end{tcolorbox}
\end{minipage}\par\medskip
\textbf{Answer evaluation.} The reference and all three edited programs pass the supplied task tests.\par
\textbf{All three edited samples.} 404, 635, 418 remaining reasoning tokens.
\end{tcolorbox}
\caption{HumanEval short-to-long prefix replacement. Bracketed ellipses indicate omissions; lengths count the full reasoning suffix after position 200. The edited continuation adds further discussion of negative inputs before producing the final program.}
\label{fig:transfer-case-humaneval-reverse}

\end{figure*}

\begin{figure*}[!p]

\begin{tcolorbox}[colback=black!1,colframe=black!25,boxrule=0.5pt,arc=1pt,left=7pt,right=7pt,top=6pt,bottom=6pt]
\small
\textbf{FOLIO long-to-short. Qwen-7B $\cdot$ QID 142 $\cdot$ $K=200$}\par\smallskip
\textbf{Task.} Premises:\newline Fort Ticonderoga is the current name for Fort Carillon.\newline Pierre de Rigaud de Vaudreuil built Fort Carillon.\newline Fort Carillon was located in New France.\newline New France is not in Europe.\par\smallskip Conclusion: Pierre de Rigaud de Vaudreuil built a fort in New France.\par\smallskip
\textbf{First-token replacement.} \textcolor{caseSource}{\textbf{\texttt{Okay}}} $\rightarrow$ \textcolor{caseEdited}{\textbf{\texttt{First}}}\par\smallskip
\textbf{Shared prefix $r_{2:200}$ (199 tokens; excerpts shown).}\par , so I need to figure out whether the conclusion is True, False, or Unknown based on the given premises. Let's break this down step by step. \textit{[\ldots]} illon. So, he built that fort.\par\smallskip Premise 3 tells us that Fort Carillon was located in New France. So, the fort is\par\medskip
\begin{minipage}[t]{0.48\linewidth}\vspace{0pt}
\begin{tcolorbox}[colback=caseSource!6,colframe=caseSource!30,boxrule=0.3pt,arc=1pt,boxsep=0pt,left=4pt,right=4pt,top=4pt,bottom=4pt,before skip=0pt,after skip=0pt]
\textcolor{caseSource}{\textbf{Original continuation}}\par\textcolor{caseSource}{\textbf{207 tokens}}\textit{; excerpts}\par\smallskip Now, the conclusion is that Pierre de Rigaud de Vaudreuil built a fort in New France. \textit{[\ldots]} \textcolor{caseSource}{\textbf{Wait, but do I need to consider anything else?}} Maybe the fact that New France is not in Europe is just additional info, but it doesn't affect the conclusion. \textit{[\ldots]} Therefore, it should be True.
\end{tcolorbox}
\end{minipage}\hfill
\begin{minipage}[t]{0.48\linewidth}\vspace{0pt}
\begin{tcolorbox}[colback=caseEdited!6,colframe=caseEdited!30,boxrule=0.3pt,arc=1pt,boxsep=0pt,left=4pt,right=4pt,top=4pt,bottom=4pt,before skip=0pt,after skip=0pt]
\textcolor{caseEdited}{\textbf{Edited continuation}}\par\textcolor{caseEdited}{\textbf{135 tokens}}\textit{; excerpts}\par\smallskip Premise 4 is just pointing out that New France is not in Europe, which might be useful context, \textcolor{caseEdited}{\textbf{but not directly related to the conclusion.}} \textit{[\ldots]} Therefore, the conclusion is definitely true based on the given premises.
\end{tcolorbox}
\end{minipage}\par\medskip
\textbf{Answer evaluation.} The reference and all three edited answers return True (gold label True).\par
\textbf{All three edited samples.} 135, 116, 217 remaining reasoning tokens.
\end{tcolorbox}
\caption{FOLIO long-to-short prefix replacement. Bracketed ellipses indicate omissions; lengths count the full reasoning suffix after position 200. The edited continuation reaches the same label with less discussion of the geographically irrelevant premise.}
\label{fig:transfer-case-folio-forward}

\par\vspace{10pt}

\begin{tcolorbox}[colback=black!1,colframe=black!25,boxrule=0.5pt,arc=1pt,left=7pt,right=7pt,top=6pt,bottom=6pt]
\small
\textbf{FOLIO short-to-long. Qwen-7B $\cdot$ QID 361 $\cdot$ $K=200$}\par\smallskip
\textbf{Task.} Premises:\newline If something is a plant, then it is not a cute animal. Simeng: All plants are not cute animals. \newline All flowers are plants.\newline Every kitten is a cute animal.\newline If something is grown in a garden, then it is a flower.\newline Piper is a kitten or a cute animal.\par\smallskip Conclusion: Piper was not grown in a garden.\par\smallskip
\textbf{First-token replacement.} \textcolor{caseSource}{\textbf{\texttt{First}}} $\rightarrow$ \textcolor{caseEdited}{\textbf{\texttt{Okay}}}\par\smallskip
\textbf{Shared prefix $r_{2:200}$ (199 tokens; excerpts shown).}\par , I'll break down the premises to understand the relationships between the entities mentioned. \textit{[\ldots]} 3 and 5, Piper must be a cute animal.\newline - From Premise 1, if something is a plant, it is not a cute\par\medskip
\begin{minipage}[t]{0.48\linewidth}\vspace{0pt}
\begin{tcolorbox}[colback=caseSource!6,colframe=caseSource!30,boxrule=0.3pt,arc=1pt,boxsep=0pt,left=4pt,right=4pt,top=4pt,bottom=4pt,before skip=0pt,after skip=0pt]
\textcolor{caseSource}{\textbf{Original continuation}}\par\textcolor{caseSource}{\textbf{85 tokens}}\textit{; complete reasoning suffix}\par\smallskip animal. Since all flowers are plants (Premise 2), flowers are not cute animals.\newline - This creates a contradiction because Piper is a cute animal but cannot be a plant or a flower. Therefore, Piper cannot be grown in a garden, as being grown in a garden implies being a flower (Premise 4), which is not cute.\par\smallskip Thus, the conclusion that Piper was not grown in a garden is true.
\end{tcolorbox}
\end{minipage}\hfill
\begin{minipage}[t]{0.48\linewidth}\vspace{0pt}
\begin{tcolorbox}[colback=caseEdited!6,colframe=caseEdited!30,boxrule=0.3pt,arc=1pt,boxsep=0pt,left=4pt,right=4pt,top=4pt,bottom=4pt,before skip=0pt,after skip=0pt]
\textcolor{caseEdited}{\textbf{Edited continuation}}\par\textcolor{caseEdited}{\textbf{92 tokens}}\textit{; complete reasoning suffix}\par\smallskip animal. Since all flowers are plants (Premise 2), any flower is not a cute animal.\newline - However, Piper is a cute animal, so she cannot be a flower or a plant because that would make her not a cute animal. Therefore, Piper cannot be a flower.\newline - From Premise 4, if something is grown in a garden, it is a flower. \textcolor{caseEdited}{\textbf{Since Piper is not a flower, she cannot be grown in a garden.}}
\end{tcolorbox}
\end{minipage}\par\medskip
\textbf{Answer evaluation.} The reference and all three edited answers return True (gold label True).\par
\textbf{All three edited samples.} 92, 88, 176 remaining reasoning tokens.
\end{tcolorbox}
\caption{FOLIO short-to-long prefix replacement. Bracketed ellipses indicate omissions; lengths count the full reasoning suffix after position 200. Both continuations derive the same label. The displayed length increase is modest; the three edited samples also show variation in its magnitude.}
\label{fig:transfer-case-folio-reverse}

\end{figure*}


\FloatBarrier
\section{Additional Sensitivity Results}
\label{app:sensitive-states}

\subsection{Concentration Across First-Token Pairs}
\begin{figure*}[!tbp]
    \centering
    \includegraphics[width=\textwidth]{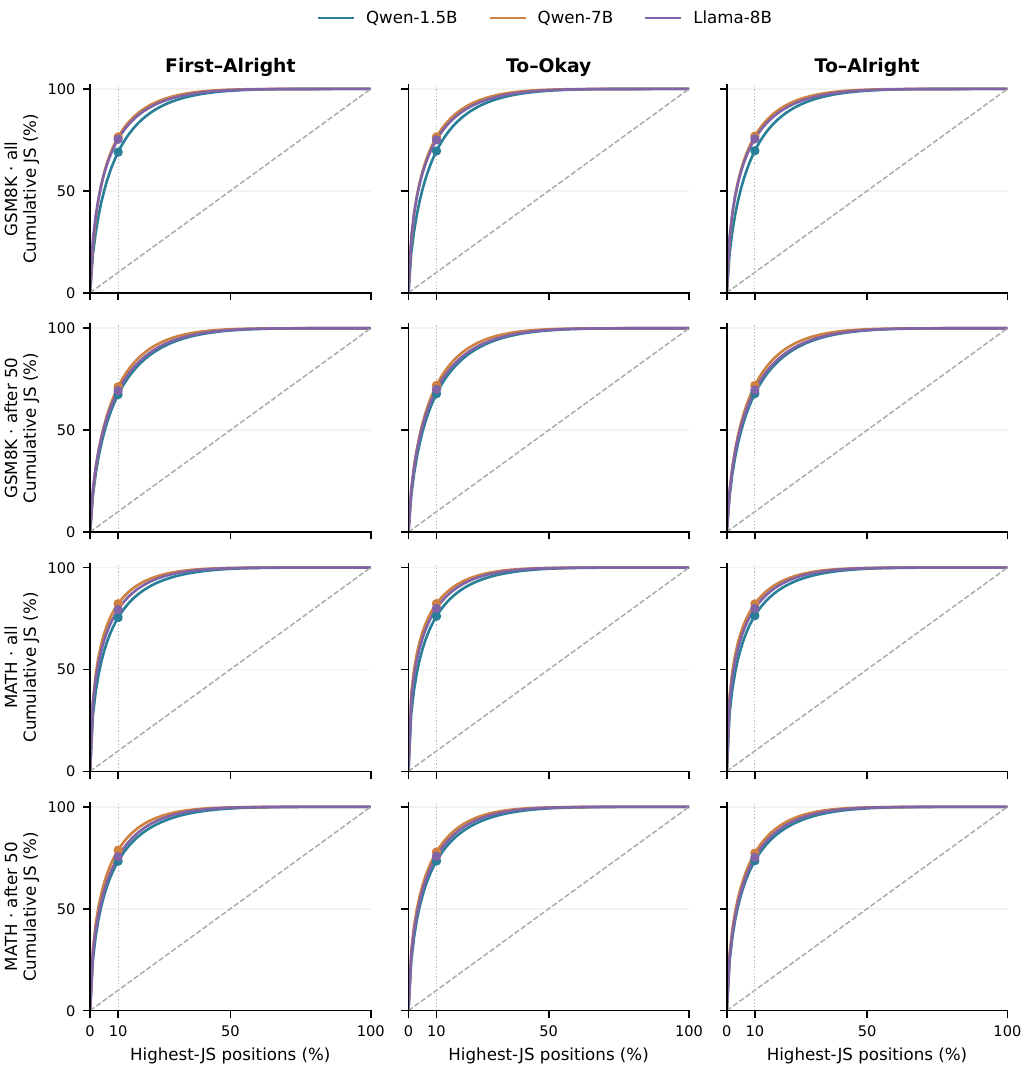}
    \caption{JS concentration for the other three long--short pairs on GSM8K and MATH. Columns identify the compared first tokens; colors identify models. For each dataset, the upper row uses all scanned positions except the first; the lower row excludes the first 50 positions and reranks and renormalizes the remainder within each trajectory. Curves average repetitions and source conditions within questions, then questions. Tail curves retain questions with available tails in all conditions and repetitions (248--250 questions per setting). Markers indicate the highest-JS 10\%; diagonals denote uniform contributions. Bands show 95\% question-bootstrap intervals from 2,000 resamples.}
    \label{fig:js-concentration-conditions}
\end{figure*}

Figure~\ref{fig:js-concentration-conditions} extends the concentration analysis to First--Alright, To--Okay, and To--Alright. 

\subsection{Attention Across First-Token Pairs}
\begin{figure*}[!tbp]
    \centering
    \includegraphics[width=\textwidth]{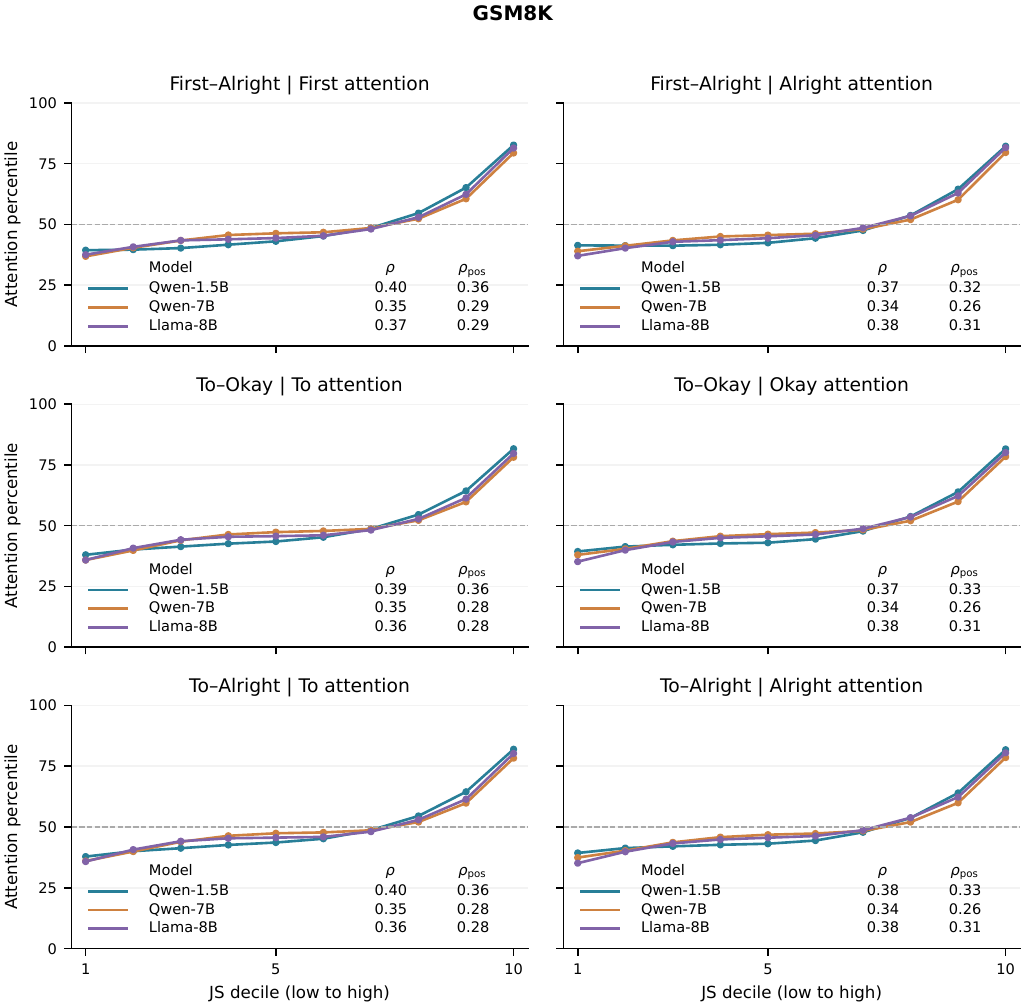}
    \caption{First-token attention versus JS for the other three long--short pairs on GSM8K. Rows identify pairs; columns show attention under the short-associated (left) and long-associated (right) first-token conditions. Curves show mean within-trajectory attention percentiles by JS decile; bands show 95\% question-bootstrap intervals. Insets report mean Spearman correlations before ($\rho$) and after ($\rho_{\mathrm{pos}}$) adjusting for token position.}
    \label{fig:attention-js-other-gsm8k}
\end{figure*}

\begin{figure*}[!tbp]
    \centering
    \includegraphics[width=\textwidth]{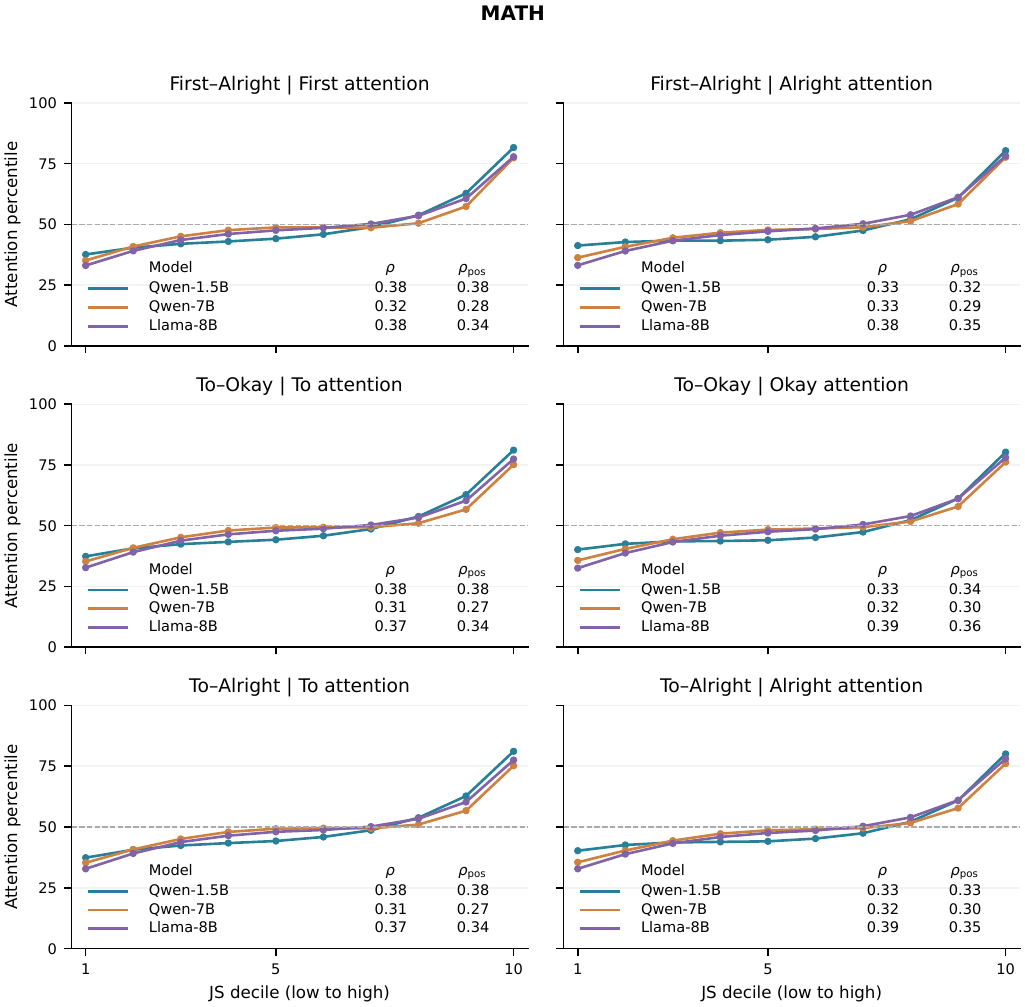}
    \caption{First-token attention versus JS for the other three long--short pairs on MATH, using the same layout and statistics as Figure~\ref{fig:attention-js-other-gsm8k}.}
    \label{fig:attention-js-other-math}
\end{figure*}

Figures~\ref{fig:attention-js-other-gsm8k} and~\ref{fig:attention-js-other-math} extend the attention analysis to the same three pairs. JS compares the paired first-token conditions on a fixed subsequent prefix; attention is measured separately under each condition. Higher-JS positions receive greater first-token attention across all three pairs.

\subsection{Sensitivity Along Individual Trajectories}
\begin{figure*}[!tbp]
    \centering
    \includegraphics[width=\textwidth]{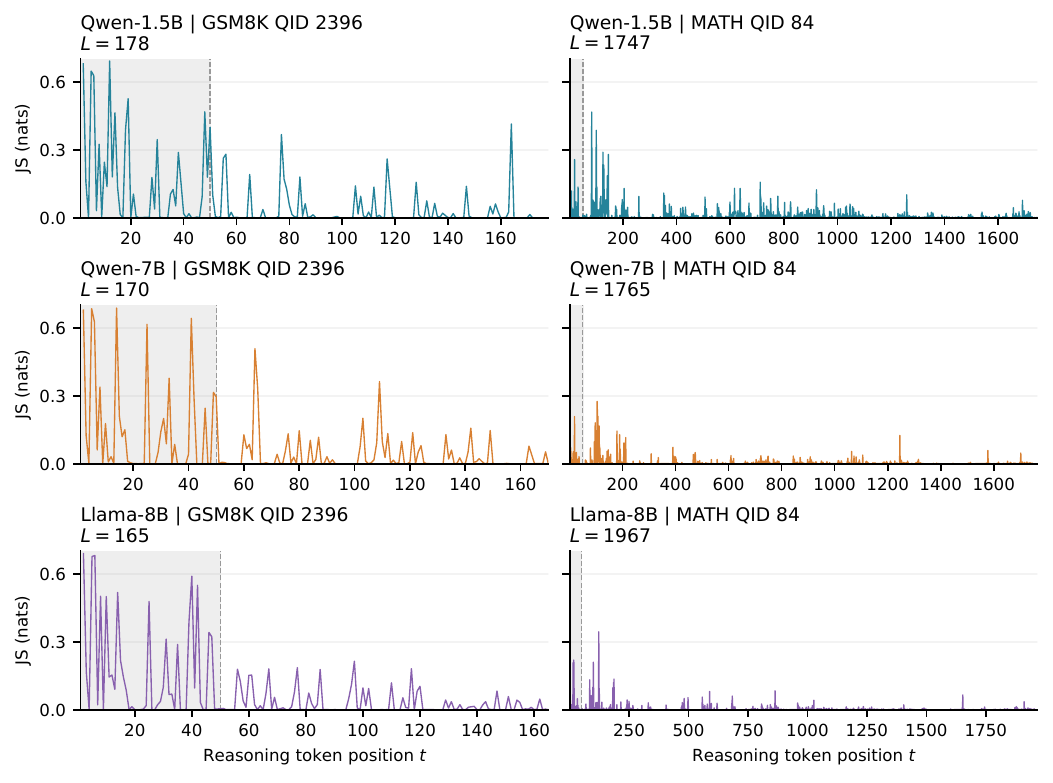}
    \caption{Unsmoothed First--Okay JS for GSM8K QID 2396 (left) and MATH-500 QID 84 (right), using natural-sampling replicate 0. QIDs are zero-based row indices in the respective dataset files; $L$ is reasoning length. Shading marks the first 50 positions. Panels show the complete scanned trajectories except the first position; horizontal scales differ with trajectory length.}
    \label{fig:js-appendix-trajectories}
\end{figure*}

Figure~\ref{fig:js-appendix-trajectories} extends the example in Figure~\ref{fig:js-concentration}(b) to three models and two datasets. The curves show intermittent peaks, including peaks beyond the initial 50 positions. For each dataset, the example is selected from complete, untruncated natural trajectories with $L>50$ in all three models, using proximity to the model-wise median lengths without inspecting JS values.

\section{Additional Intervention Results}
\label{app:interventions}


\subsection{Attention Blocking}


\begin{figure*}[!tbp]
\centering
\includegraphics[width=\textwidth]{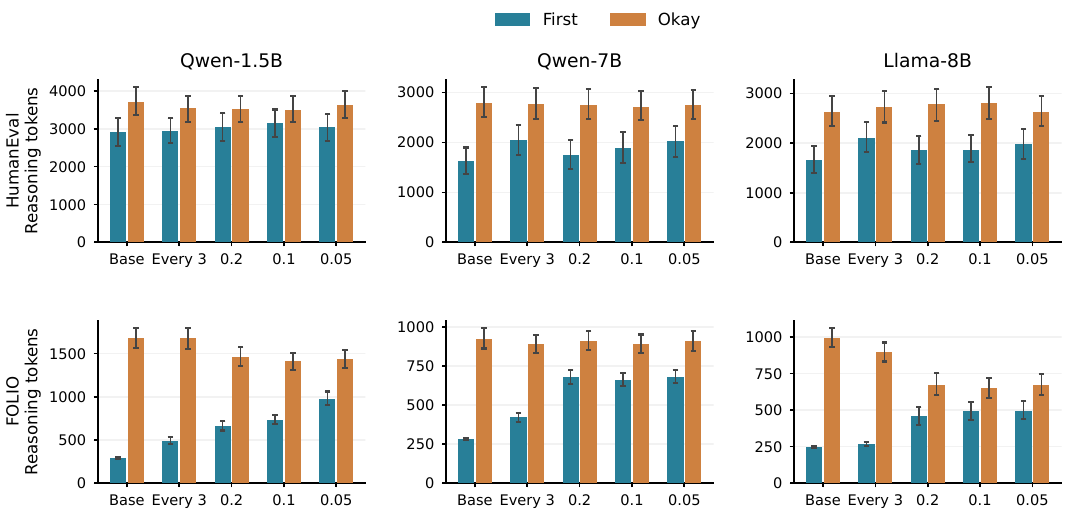}
\caption{First-token attention blocking on HumanEval (top) and FOLIO (bottom). Base denotes no intervention; Every 3 uses periodic masking; numeric headings denote JS thresholds. Error bars show 95\% question-bootstrap intervals.}
\label{fig:transfer-attention}
\par\vspace{10pt}
\begingroup
\makeatletter\def\@captype{table}\makeatother

\centering\small
\setlength{\tabcolsep}{5pt}
\begin{tabular}{llrrrrr}
\toprule
Model & Source & Base & Every 3 & $0.2$ & $0.1$ & $0.05$ \\
\midrule
\rowcolor[gray]{0.92}\multicolumn{7}{c}{HumanEval} \\
Qwen-1.5B & First & 2,910 & 2,941 (33.3) & 3,047 (0.6) & 3,145 (1.3) & 3,035 (2.7) \\
 & Okay & 3,717 & 3,536 (33.3) & 3,516 (0.2) & 3,509 (0.6) & 3,635 (1.3) \\
Qwen-7B & First & 1,622 & 2,041 (33.3) & 1,752 (2.2) & 1,889 (4.0) & 2,018 (6.5) \\
 & Okay & 2,797 & 2,763 (33.3) & 2,746 (0.3) & 2,716 (0.7) & 2,757 (1.6) \\
Llama-8B & First & 1,655 & 2,109 (33.3) & 1,853 (1.6) & 1,864 (3.3) & 1,979 (6.0) \\
 & Okay & 2,633 & 2,730 (33.3) & 2,777 (0.2) & 2,802 (0.6) & 2,629 (1.4) \\
\rowcolor[gray]{0.92}\multicolumn{7}{c}{FOLIO} \\
Qwen-1.5B & First & 291 & 494 (33.3) & 660 (1.0) & 735 (1.9) & 980 (3.4) \\
 & Okay & 1,679 & 1,676 (33.3) & 1,466 (0.5) & 1,410 (1.1) & 1,434 (2.7) \\
Qwen-7B & First & 283 & 422 (33.3) & 680 (0.9) & 663 (1.7) & 682 (3.3) \\
 & Okay & 926 & 891 (33.3) & 910 (0.6) & 891 (1.1) & 909 (2.3) \\
Llama-8B & First & 250 & 268 (33.3) & 457 (6.0) & 492 (11.1) & 497 (18.0) \\
 & Okay & 993 & 898 (33.3) & 672 (4.8) & 648 (9.5) & 671 (15.9) \\
\bottomrule
\end{tabular}
\caption{Attention blocking on transfer tasks. Cells report mean reasoning length; parentheses give mean per-trajectory intervention percentages. }
\label{tab:transfer-attention}

\endgroup
\end{figure*}

\paragraph{Extension to additional datasets.}
Figure~\ref{fig:transfer-attention} and Table~\ref{tab:transfer-attention} present the results on FOLIO and HumanEval. On FOLIO, the results follow the same pattern as on the mathematical datasets, with all tested JS thresholds reducing the length gap more than periodic masking. The results on HumanEval are less consistent: periodic masking outperforms all tested JS thresholds for Qwen-7B and Llama-8B, whereas thresholds of 0.2 and 0.1 outperform periodic masking for Qwen-1.5B. Two factors may contribute to this difference. First, the first-token length effect is weaker on HumanEval than on the other tasks, leaving less room for gap reduction. Second, the same JS thresholds select a substantially smaller fraction of positions on HumanEval, which may limit the impact of the intervention.

\paragraph{Case Study.}
Figure~\ref{fig:attention-blocking-case} compares reasoning trajectories beginning with \texttt{First} and \texttt{Okay} for the same GSM8K question. Without intervention, the \texttt{First} trajectory presents the solution and answer directly, whereas the \texttt{Okay} trajectory repeatedly checks the result, exhibiting a tendency toward overthinking. Blocking attention to the first token at sensitive positions narrows the length gap and makes the two trajectories more similar in both content and wording. This example illustrates how the intervention weakens the first-token length effect, alongside changes in reasoning behavior.

\begin{figure*}[!tp]
\centering
\input{text/attention_blocking_case}
\caption{Attention-blocking case from Qwen-7B on GSM8K (question ID 958). Columns compare First and Okay openings; rows show generation without and with attention blocking at $J_t>0.05$. The unmodified Okay trajectory is excerpted, with $[\ldots]$ marking omissions; the other reasoning traces are shown in full. All four trajectories yield the correct answer, 28.}
\label{fig:attention-blocking-case}
\end{figure*}

\subsection{Distribution Replacement}
Table~\ref{tab:distribution-intervention-rates} reports distribution-replacement rates on the mathematical datasets, using the same averaging and eligible-position denominators.
\par\smallskip\noindent\begin{minipage}{\linewidth}
\makeatletter\def\@captype{table}\makeatother
\centering\small
\renewcommand{\arraystretch}{0.82}
\setlength{\tabcolsep}{3pt}
\begin{tabular}{llrrrr}
\toprule
Model & Direction & Every 3 & $0.2$ & $0.1$ & $0.05$ \\
\midrule
\rowcolor[gray]{0.92}\multicolumn{6}{c}{GSM8K} \\
Qwen-1.5B & $F\to O$ & 33.3 & 3.4 & 5.4 & 8.5 \\
 & $O\to F$ & 33.3 & 10.6 & 16.9 & 23.7 \\
Qwen-7B & $F\to O$ & 33.3 & 3.8 & 6.0 & 9.4 \\
 & $O\to F$ & 33.4 & 7.0 & 11.2 & 16.4 \\
Llama-8B & $F\to O$ & 33.3 & 2.5 & 3.6 & 5.4 \\
 & $O\to F$ & 33.4 & 9.2 & 14.9 & 21.8 \\
\rowcolor[gray]{0.92}\multicolumn{6}{c}{MATH} \\
Qwen-1.5B & $F\to O$ & 33.3 & 1.3 & 2.3 & 4.1 \\
 & $O\to F$ & 33.3 & 5.9 & 11.0 & 17.0 \\
Qwen-7B & $F\to O$ & 33.3 & 1.3 & 2.2 & 3.7 \\
 & $O\to F$ & 33.3 & 5.3 & 9.5 & 14.2 \\
Llama-8B & $F\to O$ & 33.3 & 0.8 & 1.3 & 2.2 \\
 & $O\to F$ & 33.3 & 5.5 & 10.6 & 16.6 \\
\bottomrule
\end{tabular}
\caption{Distribution-replacement rates (\%). $F$ and $O$ denote First and Okay. Values are mean per-trajectory percentages of eligible positions; numeric headings denote JS thresholds.}
\label{tab:distribution-intervention-rates}
\end{minipage}\par\smallskip

\begin{figure*}[!tbp]
\centering
\includegraphics[width=\textwidth]{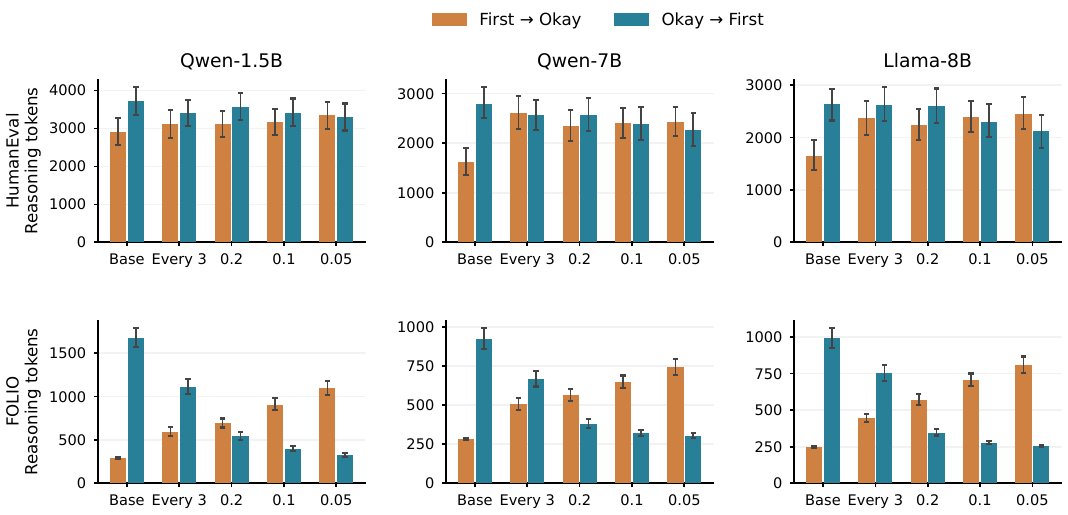}
\caption{Distribution replacement on HumanEval (top) and FOLIO (bottom). Colors identify the replacement direction; Base uses the source-opening condition, Every 3 denotes periodic replacement, and numeric headings denote JS thresholds. Error bars show 95\% question-bootstrap intervals.}
\label{fig:transfer-distribution}
\par\vspace{10pt}
\begingroup
\makeatletter\def\@captype{table}\makeatother

\centering\small
\setlength{\tabcolsep}{5pt}
\begin{tabular}{llrrrrr}
\toprule
Model & Source & Base & Every 3 & $0.2$ & $0.1$ & $0.05$ \\
\midrule
\rowcolor[gray]{0.92}\multicolumn{7}{c}{HumanEval} \\
Qwen-1.5B & First & 2,910 & 3,105 (33.3) & 3,123 (0.3) & 3,157 (0.7) & 3,343 (1.4) \\
 & Okay & 3,717 & 3,407 (33.3) & 3,574 (0.3) & 3,418 (1.0) & 3,302 (2.8) \\
Qwen-7B & First & 1,622 & 2,606 (33.3) & 2,346 (0.4) & 2,403 (0.9) & 2,425 (1.7) \\
 & Okay & 2,797 & 2,565 (33.3) & 2,572 (1.1) & 2,384 (3.3) & 2,258 (7.4) \\
Llama-8B & First & 1,655 & 2,374 (33.3) & 2,241 (0.4) & 2,396 (1.0) & 2,456 (1.9) \\
 & Okay & 2,633 & 2,621 (33.3) & 2,608 (0.7) & 2,300 (3.0) & 2,120 (7.0) \\
\rowcolor[gray]{0.92}\multicolumn{7}{c}{FOLIO} \\
Qwen-1.5B & First & 291 & 593 (33.3) & 691 (1.4) & 905 (2.0) & 1,098 (3.2) \\
 & Okay & 1,679 & 1,114 (33.3) & 541 (4.2) & 397 (8.7) & 324 (14.9) \\
Qwen-7B & First & 283 & 507 (33.3) & 567 (1.4) & 648 (2.6) & 743 (4.4) \\
 & Okay & 926 & 668 (33.3) & 381 (4.3) & 322 (8.4) & 305 (14.6) \\
Llama-8B & First & 250 & 445 (33.3) & 572 (1.8) & 705 (3.1) & 808 (5.7) \\
 & Okay & 993 & 752 (33.3) & 347 (5.3) & 277 (10.6) & 257 (18.1) \\
\bottomrule
\end{tabular}
\caption{Distribution replacement on transfer tasks. Cells report mean reasoning length; parentheses give mean per-trajectory intervention percentages. Base uses the source-opening forced trajectories. Each condition contains 492 HumanEval or 1,500 FOLIO trajectories per model.}
\label{tab:transfer-distribution}

\endgroup
\end{figure*}

\paragraph{Extension to additional datasets.}
Figure~\ref{fig:transfer-distribution} and table~\ref{tab:transfer-distribution} present the distribution replacement results on FOLIO and HumanEval. On FOLIO, the results follow the same pattern as on the mathematical datasets, with all tested JS thresholds producing larger length shifts in the intended direction than periodic replacement. On HumanEval, the relative effectiveness varies across models and replacement directions. Two factors may contribute to this variation: the first-token length effect is weaker on HumanEval, and the same JS thresholds select a substantially smaller fraction of positions, potentially limiting the impact of distribution replacement.

\end{document}